\makeatletter\def\graphicscache@inhibit{true}\makeatother
\documentclass[5p]{elsarticle}

\usepackage{lineno,hyperref}

\journal{Robotics and Autonomous Systems}

\usepackage{tikz}
\usetikzlibrary{arrows}
\usetikzlibrary{positioning,calc}
\usetikzlibrary{decorations.pathreplacing}
\usetikzlibrary{decorations.markings}
\usetikzlibrary{fit}
\usetikzlibrary{shapes.callouts}
\usetikzlibrary{shapes.geometric}
\usetikzlibrary{matrix}
\usepackage[per=frac,binary-units=true]{siunitx}

\graphicspath{{images/}}
\DeclareGraphicsExtensions{.pdf,.png,.jpg,.jpeg,.JPG}

\usepackage{bm} %
\usepackage{amsmath} %
\usepackage{amssymb}  %
\usepackage{amsfonts}

\usepackage{float}
\usepackage{makecell}
\usepackage[normalem]{ulem}

\usepackage[capitalize]{cleveref}
\usepackage{booktabs}

\usepackage{gensymb}
\usepackage[tbtags]{mathtools}
\usepackage{units}

\usepackage{textcomp}
\usepackage{siunitx}
\usepackage{multirow}

\usepackage{pgfplotstable}
\usepackage{pgfplots}
\pgfplotsset{compat=1.16}
\usepgfplotslibrary{groupplots}
\usepgfplotslibrary{units}
\usepgfplotslibrary{statistics}

\newcommand{\R}{\mathbb{R}}

\usepackage{graphicscache}

\usepackage[para,online,flushleft]{threeparttable}

\def\drawORnot{2}%
\usepackage{ifthen}

\newcommand{\drawboundingbox}[2][red]{
  \ifthenelse{\equal{#2}{1}}
    {
    \draw [#1] (current bounding box.south west) rectangle (current bounding box.north east);
    }
    {
    }%
}

\usepackage[tightpage]{preview}

\newenvironment{maybepreview}%
{\noindent\ignorespaces}%
{\par\noindent%
\ignorespacesafterend}
\PreviewEnvironment{maybepreview}

\begin{document}

\begin{frontmatter}
\title{Bimanual Telemanipulation with Force and Haptic Feedback through an Anthropomorphic Avatar System}

\author{Christian Lenz\corref{cor1}}
\ead{lenz@ais.uni-bonn.de}
\author{Sven Behnke\corref{}}
\ead{behnke@cs.uni-bonn.de}

\cortext[cor1]{Corresponding author}
\address{Institute for Computer Science VI, Autonomous Intelligent Systems, University of Bonn, Friedrich-Hirzebruch-Allee 8, 53115 Bonn, Germany}

\begin{abstract}
Robotic teleoperation is a key technology for a wide variety of applications.
It allows sending robots instead of humans in remote, possibly dangerous locations
while still using the human brain with its enormous knowledge
and creativity, especially for solving unexpected problems.
A main challenge in teleoperation consists of providing enough feedback to the human operator for situation awareness and thus create full immersion, as well as
offering the operator suitable control interfaces to achieve efficient and robust task fulfillment.
We present a bimanual telemanipulation system consisting of an anthropomorphic
avatar robot and an operator station providing force and haptic feedback to the human operator.
The avatar arms are controlled in Cartesian space with a direct mapping of the operator movements.
The measured forces and torques on the avatar side are haptically displayed to the operator.
We developed a predictive avatar model for limit avoidance which runs on the operator side, ensuring low latency.
The system was successfully evaluated during the ANA Avatar XPRIZE competition semifinals. In addition, we performed in lab experiments and
carried out a small user study with mostly untrained operators.
\end{abstract}


\end{frontmatter}

\begin{figure*}[t]
\centering\begin{maybepreview}%
 \includegraphics[height=7cm]{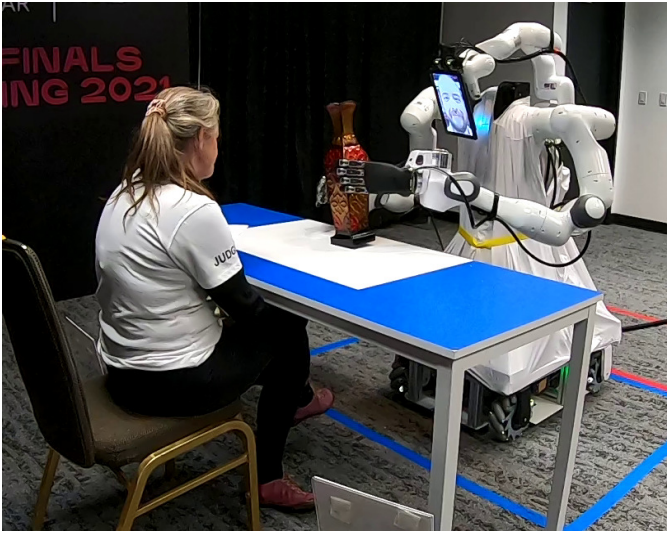}
 \hfill
 \includegraphics[height=7cm]{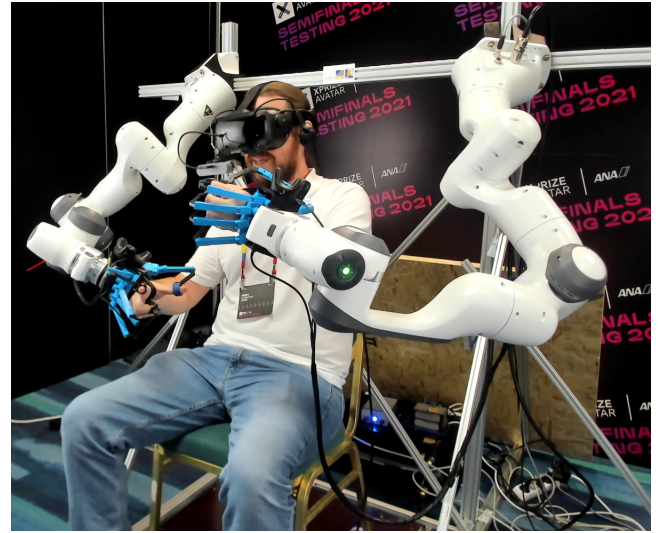}%
 \end{maybepreview}%
 \caption{Our bimanual haptic telemanipulation system: Human operator in operator station (right) inspecting an object during the ANA Avatar XPRIZE Competition
 semifinals through a remote anthropomorphic avatar robot (left).}
 \label{fig:teaser}
\end{figure*}

\section{Introduction}

\begin{tikzpicture}[remember picture,overlay]
  \node[anchor=north,align=center,font=\sffamily\small,yshift=-0.4cm] at (current page.north) {%
  \textbf{Published in Robotics and Autonomous Systems, 2022} \url{https://doi.org/10.1016/j.robot.2022.104338}
  };
\end{tikzpicture}%
Teleoperation is a very powerful method to control robots. It enables
humans to explore remote locations and to interact there with objects and persons without being physically present.
Although state-of-the-art methods for autonomous control are improving rapidly, the experience
and instincts of humans, especially for solving unpredictable problems is unparalleled so far.
The current COVID-19 pandemic is a great example of scenarios where remote work is highly desirable.
Further possible applications for teleoperation include disaster response where
humans can operate remotely and use critical situation-saving skills without risking their lives as well as
maintenance and healthcare to allow experts operating in remote locations for manipulation tasks without the need of travel.
Robotic teleoperation is a popular research area which is advanced by multiple robotic competitions like the DARPA Robotics Challenge~\citep{krotkov2017darpa} and RoboCup Rescue~\cite{kitano1999robocup}. These events are a great opportunity to
benchmark and evaluate different highly integrated and complex systems in standardized test scenarios under comparable conditions.
Our team NimbRo participates in the ANA Avatar XPRIZE Competition\footnote{\url{https://www.xprize.org/prizes/avatar}}
with the goal to advance the state of the art of such robotic telemanipulation and telepresence systems.

In addition to immersive visualization of the remote location, one important aspect is telemanipulation which enables
the operator to physically interact with the remote environment. This capability is critical for many applications---without it,
we are constrained to mere telepresence.

In this work, we present a humanoid bimanual telemanipulation system built from off-the-shelf components,
which allows a human operator to interact and manipulate in remote locations (see \cref{fig:teaser}).  Our contributions include:
\begin{enumerate}

 \item Integrating a bimanual robotic avatar and an upper-body operator exoskeleton for Cartesian telemanipulation,
 \item an arm and hand controller with force and haptic feedback,
 \item a model-based arm movement prediction to haptically display position and velocity limitations of the remote avatar in real time,
 \item an oscillation observer module to detect and suppress oscillations introduced in the force-feedback control loop, and
 \item subsystem evaluation in lab experiments, a user study, as well as our participation at the ANA Avatar XPRIZE competition semifinals.
\end{enumerate}

\section{Related Work}
\label{sec:related_work}

Teleoperation is a widely investigated research area. A leading device (in our context called the Operator Station, see \cref{sec:hardware}),
often with haptic feedback is used to control a following device (Avatar Robot) in a remote location.
The DARPA Robotics Challenge (DRC) 2015~\citep{krotkov2017darpa} required the development of mobile telemanipulation systems.
Several research groups, such as DRC-HUBO~\citep{oh2017technical}, CHIMP~\citep{stentz2015chimp}, RoboSimian~\citep{karumanchi2017team}, and our own entry Momaro~\citep{schwarz2017nimbro} presented teleoperation systems with impressive manipulation capabilities.
The focus was on completing as many manipulation and locomotion tasks as possible using a team of trained operators. Thus, some hardware and software
components were highly specialized towards solving pre-defined tasks.
In addition, the robots were not required to communicate or interact with other humans in the remote location and thus did not feature
respective capabilities. In contrast, our developed avatar solution was designed for interaction with humans in the remote location and the
operator interface is designed to give intuitive control over the robot to a single, possibly untrained operator.
Presence of
the operator in the remote location is prioritized over specific task solution skills.

Passivity control constitutes a large research field in the context of teleoperation.
Uncertainties of the operator's input dynamics, as well as the remote environment
are factors which introduce potential instability in control loops.
Many different passivity control methods tackle the stability problem, e.g.
\citep{lee2010passive, anderson1988bilateral, hannaford2002time, berestesky2004discrete, ryu2007stable, hirche2012human}.
These control schemes use the concept of passivity which is a sufficient
condition to obtain a stable control system.
Control systems are considered passive if and only if the energy flowing into
the system exceeds the energy flowing out at any time.
Conveniently, if all subsystems are passive the entire system is guaranteed to be passive as well.

\citet{ryu2007stable} use a
time-domain passivity control approach to ensure stable teleoperation,
handling time delay of up to 120\,ms. A passivity observer is used to monitor the
energy transferred from the operator to the avatar system and vice versa. The passivity
controller actively dampens the system to ensure passivity and thus stability of the system.
Our approach uses a similar observer and dampening approach to ensure stable teleoperation control loops.
One drawback of energy-based time-domain passivity controllers is the occurrence of position drift, which is
handled by~\citep{chawda2014position,artigas2010position} and improved in a 1\,DOF (degree of freedom) teleoperation setup by~\citet{coelho2018smoother}.
The design of our control architecture assures position drift-free teleoperation by commanding goal poses for each
avatar hand. Small position derivations can occur due to motion execution.
However, these are negligible small for our application (see \cref{sec:arm_controller_accuracy}).
Overall, passivity control can suffer from distortion of the displayed environment~\citep{xu2017haptic}.

In presence of large communication time delays (in the order of seconds) between operator and operator station (e.g., in space teleoperation
missions), predictive control methods can improve task performance~\citep{burridge2009using}.
The remote robot tries to anticipate human control commands to complete partially transmitted
instructions. \citet{hauser2013recognition} presents a prediction and planning method to assist teleoperation without
assuming a finite set of candidate tasks. Nevertheless, this method is limited to a finite set of task types
and needs a large number of training data to overcome this limitation.
Our control approach is designed to limit the operator only within the underlying hardware
constrains, without constraining the task solution itself.

Some recent approaches use teleoperation interfaces which only send commands to
the robot without providing any force or haptic feedback to the operator~\citep{wu2019teleoperation,rakita2017motion}.
The advantage of such systems is clearly the low weight of the capture devices which hinder the operator only marginally.
The downside is missing force or haptic feedback, especially for tasks that cannot be solved with visual feedback alone, such as difficult peg-in-hole tasks.

Other methods use custom-developed operator interfaces including force and haptic feedback. \citet{klamt2020remote}
developed a centaur-like platform for teleoperation in disaster response scenarios. The proposed teleoperation interfaces
uses actuators located at the base of the device and metallic tendons for torque transmission to the actuated joints.
This approach benefits from light-weight moving parts with low inertia resulting in an easily back-drivable system.
However, the used metallic tendons introduce some compliant behavior which need to be considered. In our approach,
we use off-the-shelf robotic arms with actuators located inside the joints. We utilize an external FT-Sensor
to actively follow the operator's arm movement. \citet{guanyang2019haptic} use two haptic devices (3DoF rotational and
3DoF translational device) to control a robot's end-effector. Both devices can display contact forces with high stiffness due to mechanical design. However, limited workspace and thus the requirement of a motion mapping between
operator station and controlled robot as well as using two devices to control full 6D motion are some drawbacks
compared to our approach.
In \citet{abi2018humanoid} the operator station is comparable to our approach, but the focus there is placed on haptic feedback for balance control of the bipedal humanoid avatar robot.

Wearable haptic feedback devices~\citep{bimbo2017teleoperation} overcome the workspace constraints generated by stationary devices but are limited to displaying contact since they cannot create any force towards the operator.
Other research projects focus on controlling a teleoperation system under time delays~\citep{niemeyer2004telemanipulation} or with two different kinematic chains on the avatar side~\citep{porcini2020evaluation}.

In contrast to the highlighted related research, our approach focuses on off-the-shelf components which allow
for easy replication and maintenance. Furthermore, the used robotic arms are replaceable with any other appropriate
actuators with different kinematic chains, since the whole communication between the systems uses only the 6D end-effector pose.

\section{Hardware Setup}
\label{sec:hardware}

\begin{figure}[]
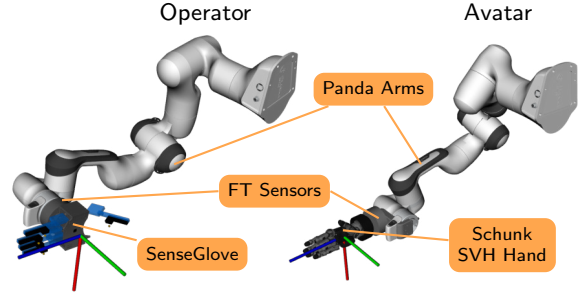

\centering\begin{maybepreview}%
\begin{tikzpicture}[
 	font=\sffamily\scriptsize,
    every node/.append style={text depth=.2ex},
	box/.style={rectangle, inner sep=0.2, anchor=west, align=center},
	line/.style={black, thick}
]
\tikzset{every node/.append style={node distance=3.0cm}}
\tikzset{content_node/.append style={minimum size=1.5em,minimum height=3em,minimum width={width("Search Point")+0.2em},draw,align=center,fill=blue!15!white, rounded corners}}
\tikzset{header_node/.append style={minimum size=1.5em,minimum height=3em,minimum width={width("Search Point")+0.2em},align=center, rounded corners}}
\tikzset{label_node/.append style={near start}}
\tikzset{group_node/.append style={align=center,rounded corners,draw, dashed , inner sep=1em,thick}}

\node (otto)[] at (0.0,0.0) {\includegraphics[height=3.7cm]{images/otto_arm.png}};
\node (anna)[] at (3.7,0.0) {\includegraphics[height=3.3cm]{images/anna_arm.png}};

\begin{scope}[x={($ (otto.south east) - (otto.south west) $ )},y={( $ (otto.north west) - (otto.south west)$ )}, shift={(otto.south west)}]
\coordinate(otto_ft) at (0.24,0.38);
   \node[fill=orange!70,rounded corners, align=center] (senseGlove) at (0.6,0.2) {SenseGlove};
   \draw[draw=orange!70, thick] (senseGlove) -- (0.27,0.3);

   \node[fill=orange!70,rounded corners, align=center] (panda) at (1.1,0.75) {Panda Arms};
   \draw[draw=orange!70, thick] (panda) -- (0.55,0.5);

  \end{scope}

\begin{scope}[x={($ (anna.south east) - (anna.south west) $ )},y={( $ (anna.north west) - (anna.south west)$ )}, shift={(anna.south west)}]
   \node[fill=orange!70,rounded corners, align=center] (ft) at (0.0,0.4) {FT Sensors};
   \coordinate(anna_ft) at (0.32,0.3);
   \coordinate(anna_panda) at (0.45,0.5);
   \node[fill=orange!70,rounded corners, align=center] (schunk) at (0.7,0.2) {Schunk\\SVH Hand};
   \draw[draw=orange!70, thick] (schunk) -- (0.21,0.255);
  \end{scope}

  \draw[draw=orange!70, thick] (anna_ft) -- (ft.east);
   \draw[draw=orange!70, thick] (otto_ft) -- (ft.west);
   \draw[draw=orange!70, thick] (panda) -- (anna_panda);

   \node at(0.65,2) {\small Operator};
   \node at(4.5,2) {\small Avatar};
\end{tikzpicture}
\end{maybepreview}%
 \caption{Operator (left) and avatar (right) arm with used hardware components. For simplicity, only the right arm is shown. The axes depict the common hand frame which is used for control commands and feedback.}
 \label{fig:arms}
\end{figure}

\begin{figure*}
\centering\begin{maybepreview}%
\begin{tikzpicture}[
 	font=\sffamily\small,
    every node/.append style={text depth=.2ex},
	box/.style={rectangle, inner sep=0.2, anchor=west, align=center},
	line/.style={black, thick}
]
\tikzset{every node/.append style={node distance=2.0cm}}
\tikzset{terminal_node/.append style={minimum size=1.0em,minimum height=3em,minimum width=3em,draw,align=center,rounded corners,execute at begin node=\setlength{\baselineskip}{2.5ex}}}
\tikzset{content_node/.append style={minimum size=1.5em,minimum height=3em,minimum width={width("Search Point")+0.2em},draw,align=center,fill=blue!15!white, rounded corners}}
\tikzset{header_node/.append style={minimum size=1.5em,minimum height=3em,minimum width={width("Search Point")+0.2em},align=center, rounded corners}}
\tikzset{label_node/.append style={near start}}
\tikzset{group_node/.append style={align=center,rounded corners,draw, dashed , inner sep=1em,thick}}
\tikzset{decision_node/.append style={align=center,shape aspect=1.5,minimum width=7.9em,minimum height=5.4em,diamond,draw,fill=yellow!25!white,font=\sffamily\normalsize,node distance=3.9cm}}

\node (operator)[] at (-2.3,0) {\includegraphics[height=4.3cm]{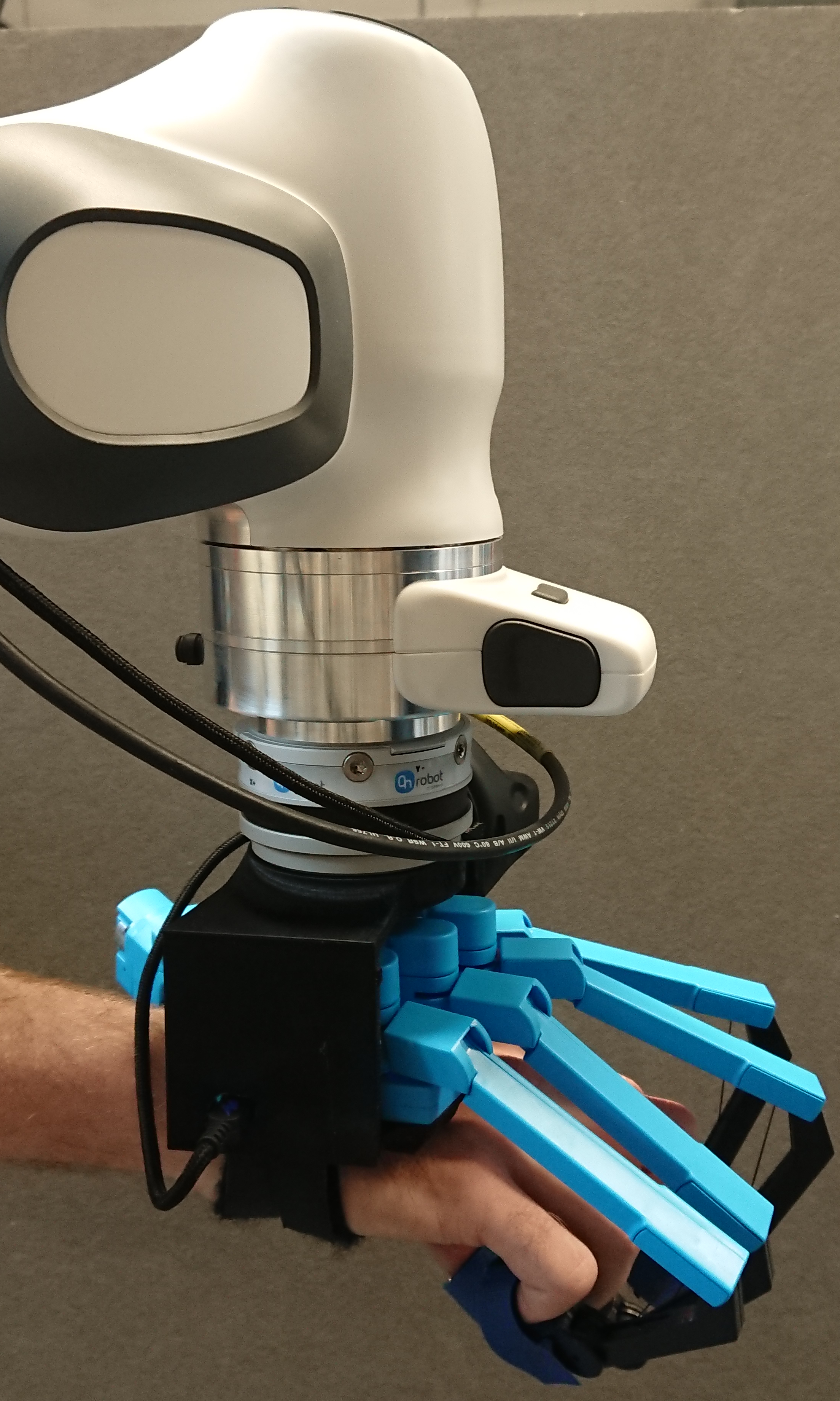}};
\node (avatar)[] at (12.6,0.0) {\includegraphics[height=4.3cm]{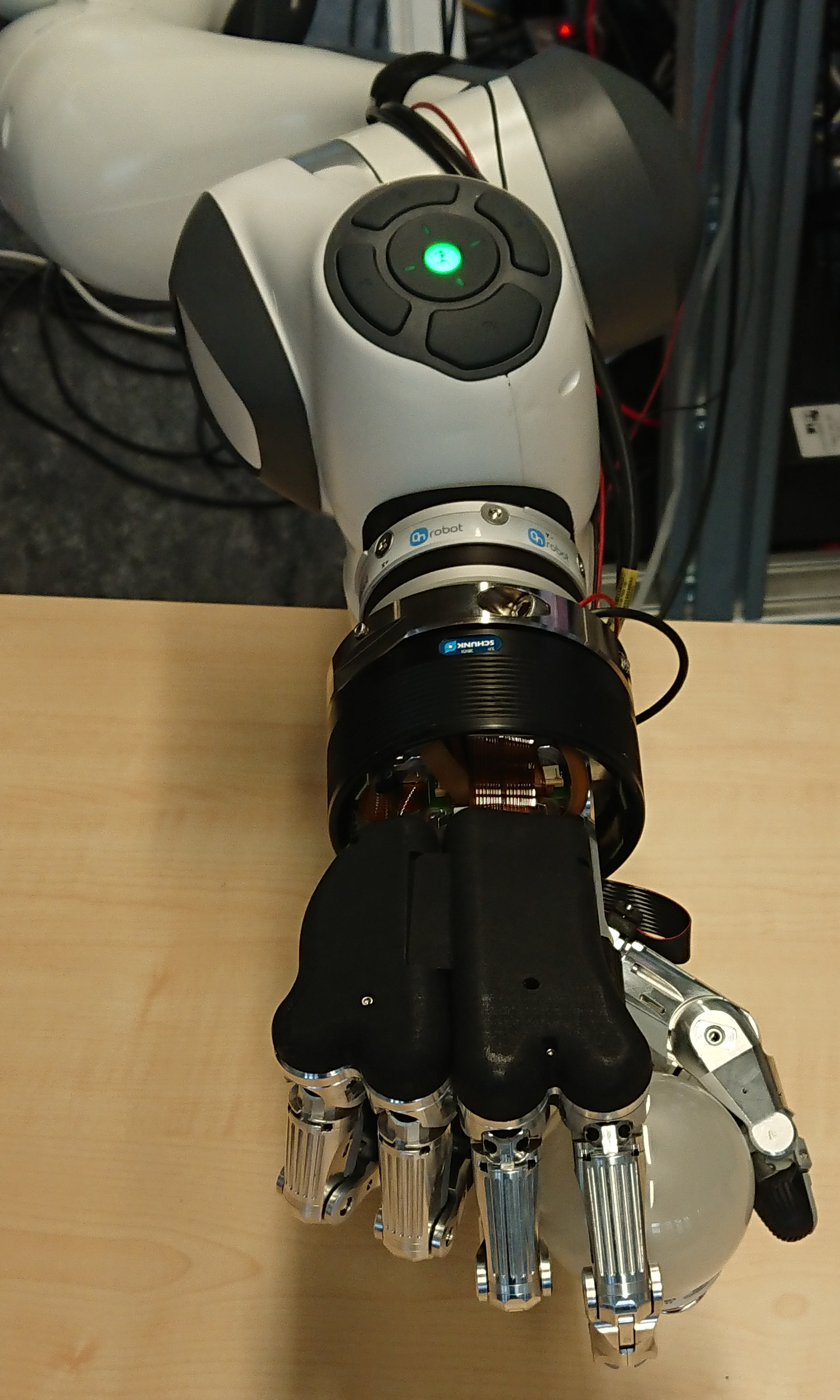}};

\node(ottoFranka)[terminal_node] at (-0.2, 1.6){Franka\\Arm};
\node(ottoFT)[terminal_node] at(-0.2,0.0) {FT\\Sensor};
\node(senseGlove)[terminal_node] at(-0.2,-1.6) {Sense\\Glove};

\node(otto_arm_controller) [terminal_node] at(2.65, 1.6) {Arm\\Controller};
\node(otto_hand_controller) [terminal_node] at(2.65, -1.6) {Hand\\Controller};

 \draw[->]([yshift = 0.2cm] ottoFranka.east)--([yshift = 0.2cm]otto_arm_controller.west) node[midway, above]{\fontsize{6pt}{6pt}\selectfont 6D pose};
 \draw[<-]([yshift = -0.2cm] ottoFranka.east)--([yshift = -0.2cm]otto_arm_controller.west) node[midway, below]{\fontsize{6pt}{6pt}\selectfont7DoF torque};

\node(anna_model) [terminal_node] at(4.8, 0.8) {Avatar\\Model};
 \draw[->](ottoFT.east) -| node[below, pos=0.25]{\fontsize{6pt}{6pt}\selectfont6D forces/torques}([xshift = -0.3cm]otto_arm_controller.south);
 \draw[->](anna_model.west) -| node[below, pos=0.25]{\fontsize{6pt}{6pt}\selectfont7D torque}([xshift = 0.3cm]otto_arm_controller.south);

 \draw [->] ([yshift = 0.2cm]senseGlove.east) -- ([yshift = 0.2cm]otto_hand_controller.west) node[midway, above]{\fontsize{6pt}{6pt}\selectfont Finger pos.};
\draw [<-] ([yshift = -0.2cm]senseGlove.east) -- ([yshift = -0.2cm]otto_hand_controller.west) node[midway, below]{\fontsize{6pt}{6pt}\selectfont Finger forces};

\node(anna_arm_controller) [terminal_node] at(7.4, 1.6) {Arm\\Controller};
\node(anna_hand_controller) [terminal_node] at(7.4, -1.6) {Hand\\Controller};

\node(annaFranka) [terminal_node] at (10.5,1.6){Franka\\Arm};
\node(annaFT) [terminal_node] at (10.5,0.0){FT\\Sensor};
\node(schunk) [terminal_node] at (10.5,-1.6){Schunk\\Hand};

\draw [->] ([yshift = 0.2cm]anna_hand_controller.east) -- ([yshift = 0.2cm]schunk.west) node[midway, above]{\fontsize{6pt}{6pt}\selectfont Finger cmds};
\draw [<-] ([yshift = -0.2cm]anna_hand_controller.east) -- ([yshift = -0.2cm]schunk.west) node[midway, below]{\fontsize{6pt}{6pt}\selectfont Motor currents};

\draw [->] ([yshift = 0.2cm]otto_hand_controller.east) -- ([yshift = 0.2cm]anna_hand_controller.west) node[above, pos= 0.35]{\fontsize{6pt}{6pt}\selectfont Hand command};
\draw [<-] ([yshift = -0.2cm]otto_hand_controller.east) -- ([yshift = -0.2cm]anna_hand_controller.west) node[below, pos= 0.35]{\fontsize{6pt}{6pt}\selectfont Hand feedback};

\draw [->] (otto_arm_controller.east) -- (anna_arm_controller.west) node[above, pos=0.35] {\fontsize{6pt}{6pt}\selectfont 6D pose};
\draw [->] (otto_arm_controller.east) -| (anna_model.north);
\draw [<-] (anna_model.east) -| (anna_arm_controller.south) node [below, pos = 0.4] {\fontsize{6pt}{6pt}\selectfont 7DoF positions};

\draw [->] ([yshift = 0.2cm]anna_arm_controller.east) -- ([yshift=0.2cm]annaFranka.west) node [above, midway] {\fontsize{6pt}{6pt}\selectfont 7DoF torques};
\draw [<-] ([yshift = -0.2cm]anna_arm_controller.east) -- ([yshift=-0.2cm]annaFranka.west) node [below, midway] {\fontsize{6pt}{6pt}\selectfont 7DoF positions};

\draw [<-] (otto_arm_controller.south) |- (annaFT.west) node[below, pos= 0.9] {\fontsize{6pt}{6pt}\selectfont 6D forces/torques};

\def\top{3.4}
\def\leftmid{5.6}
\def\rightmid{6.05}
\def\left{-3.8}
\def\right{14.1}
\draw [dashed, gray] (\left, -2.5) -- (\left, \top) -- (\leftmid, \top) -- (\leftmid, -2.5) -- (\left, -2.5);
\node()[header_node, align = center] at (1.0, 2.8) {\large Operator Station};

\draw [dashed, gray] (\rightmid, -2.5) -- (\rightmid, \top) -- (\right, \top) -- (\right, -2.5) -- (\rightmid, -2.5);
\node()[header_node, align = center] at (9.87, 2.8) {\large Avatar Robot};

\end{tikzpicture}\end{maybepreview}%
\caption{Control System overview. For simplicity, only the right side is depicted. The left side is controlled similarly, besides a different Hand Controller for the different Schunk hands.
}
\label{fig:system}

\end{figure*}
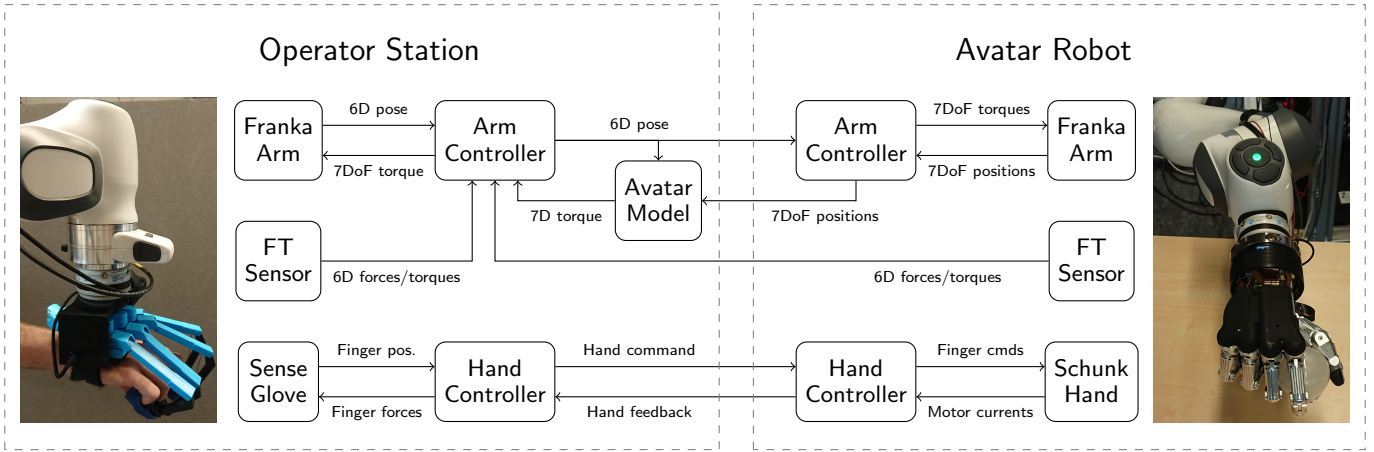

The developed robotic teleoperation system consists of an operator station and an avatar robot, as shown in \cref{fig:teaser}.
The operator station allows the operator to control the avatar from a remote location. It includes two robotic arms,
hand exoskeleton, 3D Rudder foot paddle, and a head mounted display with additional sensors.
The avatar robot is designed to interact with humans and in human-made indoor environments and, thus,
features an anthropomorphic upper body mounted on a mobile base.

The operator station and the avatar robot are controlled with a standard desktop
computer (Intel i9-9900K @ 3.60\,GHz, NVidia RTX 2080) each.
The communication between these computers is achieved by a single Gigabit
Ethernet connection. We successfully tested the system with artificial delay of up to 30\,ms in
both directions. Thus, our system allows operating the avatar from a distant location.
On the software side, the Robot Operating System (ROS) framework is used.
Both, the operator station and the avatar robot run their own roscore. We use NimbRo Network\footnote{\url{https://github.com/AIS-Bonn/nimbro_network}}
for any communication between both roscores.
The hardware design of the operator station and avatar robot is described in the following.

\subsection{Avatar Robot}

The avatar robot's anthropomorphic upper body mimics the human arm configuration
using two 7\,DoF Franka Emika Panda arms, mounted in slightly V-shaped
angle. The shoulder height of 110\,cm above
the floor allows convenient manipulation of objects on a table, as well
interaction with both sitting and standing persons. The shoulder width
of under 90\,cm enables easy navigation through standard doors.

The Panda arms have a sufficient payload of 3\,kg and a maximal reach of
855\,mm. The extra degree of freedom gives some flexibility in the
elbow position. While the arm measures joint torques
in each arm joint, we mounted additional OnRobot HEX-E 6-Axis force/torque
sensors at the wrists for more accurate force and torque measurements
close to the robotic hands, since this is the default location of contact
with the robot's environment (see \cref{fig:arms}). The avatar robot is equipped with two anthropomorphic hands.
A 20\,DoF Schunk SVH hand is mounted on the right side. The nine actuated
DoF provide very dexterous manipulation capabilities.
The left arm features a 5\,DoF Schunk SIH hand for simpler but more
force-requiring manipulation tasks. Both hand types thus complement each other.

The avatar's head is equipped with two RGB cameras, a microphone,
and a small screen displaying the animated face of the operator~\citep{rochow2022vr}. It is attached to the
upper body using a 6\,DoF UFactory xArm for free head movement.
In addition, two wide-angle RGB cameras are capturing the robot's vicinity for situation awareness during locomotion.
Further details on the VR remote visualization system are provided in~\citep{schwarz2021humanoid}.
The anthropomorphic upper body has been mounted on a movable base, which allows omnidirectional movement.

\subsection{Operator Station}

The operator controls the avatar through the Operator Station from a
comfortable sitting pose. The human hand movement is captured with a similar
setup as already described for the avatar robot: Two Panda arms
are equipped with an OnRobot HEX-E force/torque sensor and connected to the
operator hand using a SenseGlove haptic interaction device. The Panda arms thus
serve dual purposes: They provide precise 6D human hand pose measurements for avatar
control, as well as the possibility to induce force feedback measured by the
Avatar onto the human wrists. The operator-side force/torque sensor is used to measure the slightest
operator hand movement to
assist the operator in moving their arm, reducing the felt mass and friction to
a minimum.

The SenseGlove haptic interaction device features 20\,DoF finger joint
position measurements (four per finger) and a 1\,DoF haptic feedback channel
per finger (i.e., when activated the human feels resistance, which prevents
further finger closing movement).

For visual and audio communication, the operator is wearing a head mounted
display equipped with eye trackers, audio headset, and a camera viewing the
lower face part (for more details see~\citep{avatar2021iros} and~\citep{schwarz2021low}). The avatar locomotion can be controlled using a 3D Rudder
foot paddle device.

The Panda arms feature built-in safety measures and will stop immediately
if force, torque, or velocity limits are exceeded. This ensures safe human-robot
interactions both on the operator and the avatar side.

\section{Force Feedback Controller}

The control architecture for the force feedback teleoperation system consists
of two arm and two hand controllers (one for each the operator and the avatar side). For
the right and the left arm, each controller pair is running separately. The hand controller
for the right and left hand are slightly different since different robotic hands are used.
An overview of the control architecture is shown in \cref{fig:system}.
The arm controllers run with an update rate of 1\,kHz and the force-torque sensor
measurements are captured with 500\,Hz. The force-torque measurements are smoothed
using a sensor-sided low-pass filter with a cut-of frequency of 15\,Hz.

Since the robot arms are attached from outside to the operator's wrists (see \cref{fig:system}),
the kinematic chains of avatar and operator station differ,
and thus, a joint-by-joint mapping of the operator
and avatar arm is not possible. Consequently, the developed control concept does not rely on similar
kinematic chains. Instead, a common control frame is defined in the middle of
the palm of both the human and robotic hands, i.e., all necessary command and
feedback data are transformed such that they refer to this frame before being
transmitted.
The controllers for the operator and avatar arms and both hands are described
in the following subsections.

\subsection{Operator Arm Controller}
\label{sec:control}
The operator arm controller commands joint torques to the Panda arm and
reads the current operator hand pose to generate the commanded hand pose sent to
the avatar robot. The goal is to generate a weightless feeling for the
operator while moving the arm---if no force feedback is displayed.
Even though the Panda arm has a convenient teach-mode using the internal
gravity compensation when zero torques are commanded, the weightless feeling
can be further improved by using precise external force-torque measurements.
Any contact established by the avatar robot with its hands and lower arms is
haptically displayed to the operator. Since the teleoperation system has no
information about the operator's intention, contact should not be avoided but
displayed to the operator to keep the human in control of the situation.

For simplicity, just one arm is mentioned in the following, since the left
and right arms are controlled equally.

\subsubsection{Torque Controller}
\label{sec:operator_arm_control}

\begin{figure}
\centering\begin{maybepreview}%
 \includegraphics[width=.95\linewidth]{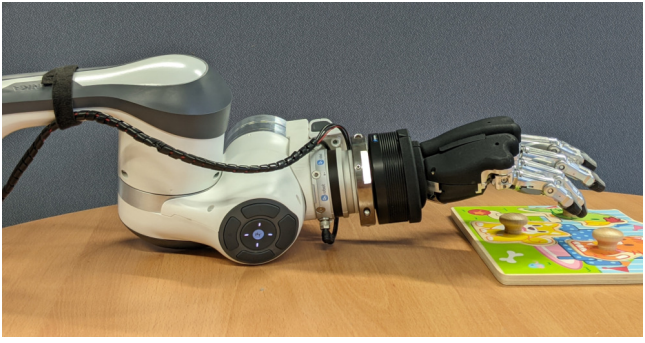}
 \end{maybepreview}%
 \caption{Unintended lower arm contact: Typical situation in which the avatar's lower arm establishes contact with the environment, which cannot be measured by the force-torque sensor.
 Panda arm torques are used to provide the operator with appropriate force feedback.}
 \label{fig:wrist_collision}
\end{figure}

\begin{figure}[t]
  \centering\begin{maybepreview}%
  \begin{tikzpicture}[font=\footnotesize\sffamily]
  \centering
  \begin{axis}[height=4cm,width=.95\linewidth,
     xlabel={$d_p$ [rad]}, ylabel={$\alpha$},
     xmin=0,xmax=5,ymin=-0.1,ymax=1.1,
     xticklabels={,,$t_p$,,$\frac{1}{2} t_p$,,0},
     xtick pos=left,
     ]
     \draw [black] (0.0, 1.0) -- (1.0, 1.0);
     \draw [black] (1.0, 1.0) -- (3, 0.0);
	 \draw [black] (3.0, 0.0) -- (5, 0.0);
   \end{axis}
  \end{tikzpicture}\end{maybepreview}%
  \caption{Operator arm torque command (see \cref{eq:operator_control}) is scaled using $\alpha$ to reduce oscillations when getting close to joint position or velocity limits. The scalar decreases linearly if the distance to a joint position limit $d_p$ exceeds the threshold $t_p$. Velocity limits are handled analogously.}
  \label{fig:alpha}
\end{figure}
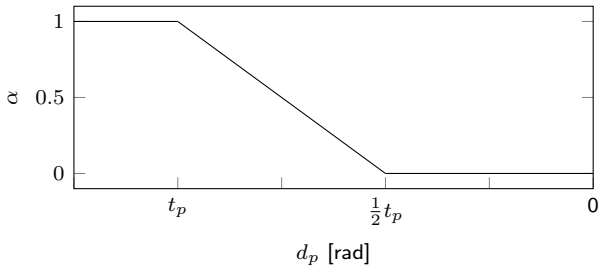

Let us denote with $\tau_o \in \R^7$ the commanded joint torques for a particular time step. Then
\begin{equation}
 \tau_o = \alpha \tau_{cmd} + \beta \tau_{f} + \tau_{lo} + \tau_{la} + \tau_{no} + \tau_{co}\label{eq:operator_control}
\end{equation}
describes the used torque components (command, force feedback, operator limit avoidance, avatar limit avoidance, null-space, and Coriolis) which will be explained in the following.
Note that the gravity compensation is not considered here, since it is
done by the Franka Control Interface (FCI) itself.

The commanded joint torques $\tau_{cmd}$ to move the Panda arm based on the
force/torque sensor measurements and are defined as

\begin{equation}
\tau_{cmd} = J^T F
\label{m:tau_cmd}
 \end{equation}
with $J$ being the body Jacobian relative to the hand frame and $F\in\R^6$ denoting the
measured 3D forces and 3D torques. The scalars $\alpha \in \R^7$ are computed by the predictive limit avoidance module (see \cref{sec:limit_avoidance}).
Note that $F$ has to be corrected taking sensor bias and attached end-effector
weight into account, as well as transformed into the common hand control frame (see \cref{sec:ft_calibration}).

The term $\tau_f$ denotes the force feedback induced by the avatar-side force-torque
sensor and Panda arm torque measurements. The scalar $\beta$ is computed by the oscillation observer module (see \cref{sec:observer}) to prevent possible oscillations in
the feedback loop.
The force-torque measurements are used as the primary feedback source. They are already bias-corrected and correctly transformed, therefore \cref{m:tau_cmd} can be directly applied analogously to compute the induced joint torques $\tau_{sensor}$.

In some situations, especially when performing manipulation tasks on a table, the avatar
establishes contact between the lower arm and the environment
(for example the table, see \cref{fig:wrist_collision}), which are
not visible from the operator's view pose. Since the force-torque sensor and
hand are above the table, this type of contact cannot be measured by the force-torque
sensor. Here, we must use the joint torque measurements of the Panda arm to give the
operator feedback about the contact.
The Franka API provides estimated Cartesian forces $f_{panda}$ at the end-effector.
We use
\begin{equation}
f_{diff} = \hat{f}_{panda} - \hat{f}_{sensor}
\end{equation}
to calculate the forces in the end-effector frame which cannot be measured using the force-torque sensor, where $\hat{f}_{panda}$ and $\hat{f}_{sensor}$ are
the respective low-pass filtered end-effector forces. Finally, we transform the calculated forces into the wrist frame and compute
\begin{equation}
 \tau_f = \tau_{sensor} + \tau_{diff},
\end{equation}
where $\tau_{diff} \in \R^6$ is $f_{diff} \in \R^3$ extended with $t_0 \in \R^3$, since we ignore torque measurements here.

\begin{figure*}[]
  \centering\begin{maybepreview}%

\pgfplotstableread{exp/dtf_off_crop.txt}\dfttable

  \begin{tikzpicture}[font=\footnotesize\sffamily]
   \begin{axis}[height=6cm,width=.75\linewidth,
     xlabel={Time [s]}, ylabel={Amplitude [N]},
     ymax=3.2,
     legend pos = north east,
     legend style={nodes={scale=0.7, transform shape}},
     ]
     \addplot+ [mark=none, red, each nth point=1, filter discard warning=false, unbounded coords=discard] table [x=t,y expr=\thisrowno{1} / 512] {\dfttable};
     \addplot+ [mark=none, green, each nth point=1, filter discard warning=false, unbounded coords=discard] table [x=t,y expr=\thisrowno{2} / 512] {\dfttable};
     \addplot+ [mark=none, blue, each nth point=1, filter discard warning=false, unbounded coords=discard] table [x=t,y expr=\thisrowno{3} / 512] {\dfttable};
     \addlegendentry{x-Axis}
     \addlegendentry{y-Axis}
     \addlegendentry{z-Axis}
     \draw [black] (0.5, -1) -- (0.5, 1.8);
     \node[align=center, anchor=east, execute at begin node=\setlength{\baselineskip}{2.5ex}] at (0.55,1.5) {Initial\\Contact};
   \end{axis}
   \node (operator)[] at (13.8,1.94) {\includegraphics[height=5cm]{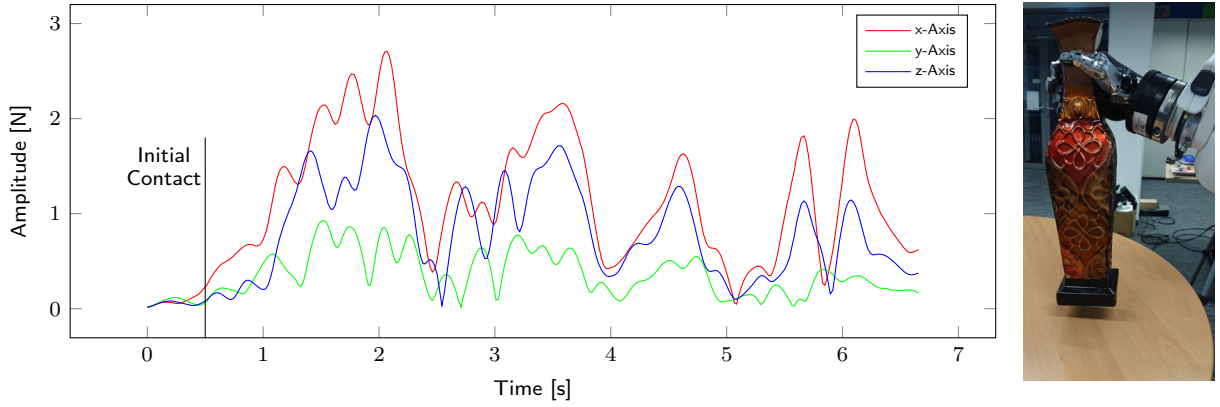}};
  \end{tikzpicture}
   \end{maybepreview}%

  \caption{Oscillation observer: Frequency analysis (left) of force feedback measurements from the avatar robot's right arm while
  placing a vase on a table (right). The initial contact between vase and table is marked. The graph shows the amplitude of the 4th frequency (ca. 5.7\,Hz) computed using a sliding DFT for each axis over time.}
  \label{fig:dft_explain}

\end{figure*}

\label{sec:avatar_arm_control}

\begin{figure}
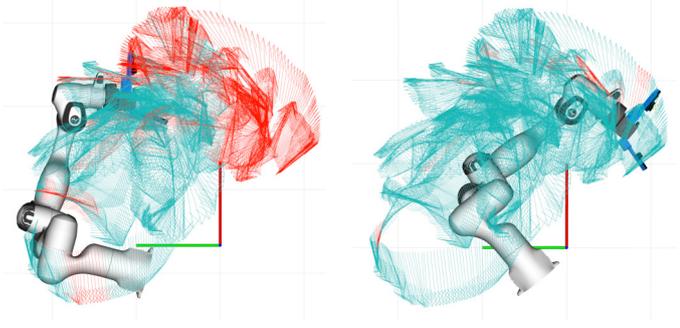

\begin{maybepreview}%
 \includegraphics[width=.48\linewidth]{images/experiments/now_final.png}\hfill
 \includegraphics[width=.48\linewidth]{images/experiments/new_final.png}
\end{maybepreview}%
 \caption{Arm workspace evaluation. Left: Initial arm setup, similar to the avatar side. Right: Optimized mounting pose. Turquoise (reachable) and red (not reachable) arrows depict the captured human left-hand poses. The coordinate axes depict the operator sitting pose.}
 \label{fig:exp_arm_reach}
\end{figure}

\subsubsection{Predictive Limit Avoidance}
\label{sec:limit_avoidance}
Humans can achieve high speeds moving their arm, which can exceed the Panda joint velocity limits (up to $150^\circ / s$).
To prevent the operator from exceeding joint position or velocity limits
of the Panda arm, the term $\tau_{lo}\in\R^7$ is introduced to apply torques
pushing the arm away from those limits. For a single joint $i$, the torque to
avoid its position limit is defined as

\begin{alignat}{2}
 \tau^i_{lo-position} = \left\{\begin{array}{ll} \gamma_p (\frac{1}{d^i_p} - \frac{1}{t_p}), & d^i_p < t_p \\
         0, & \text{else}\end{array}\right.
\end{alignat}
with $\gamma_p$ being a constant scalar, $d^i_p$ being the distance for
joint $i$ to its closer position limit, and $t_p = 10^\circ$ being a threshold
how close a joint must be at a limit to activate this behavior. $\tau_{lo-velocity}^i$
is calculated analogously with $t_v = 40^\circ / sec$. Together, $\tau_{lo}$ is defined as
\begin{equation}
 \tau_{lo} = \tau_{lo-position} + \tau_{lo-velocity}. \label{eq:avoidance}
\end{equation}
The torques $\tau_{lo}$ exhibit hyperbolical growth when getting closer to respective limits.
Since the operator-side force-torque sensor will measure the generated limit
avoidance torques, the arm can end up oscillating, especially being close to one
or multiple position limits. Thus, the torques $\tau_{cmd}$, which are
influenced by the force-torque sensor, are scaled per joint by $\alpha$, which is defined as

\begin{equation}
\alpha = \max(0, \min(1, 2 \min(\frac{d_p}{t_p}, \frac{d_v}{t_v}) - 1)).
\end{equation}
The scalar $\alpha$ is designed to decrease linearly and reach zero when the limit is approached halfway after activating the limit avoidance (see \cref{fig:alpha}). This reduces the commanded torques $\tau_{cmd}$ enough when approaching a position or velocity limit and prevents the oscillation.

As already mentioned, the operator station and avatar robot have different kinematic arm chains. Therefore, avoiding position and velocity limits on the operator side does not guarantee limit avoidance on the avatar side. Calculating the joint torques preventing joint limits on the avatar robot in a similar way is not beneficial, since the feedback information would arrive with high latency (mainly because of the delay generated by motion execution on the avatar side).
To overcome this issue, we use a model of the avatar inside the operator arm controller to predict the avatar arm movement for the next time step and calculate the needed joint torques to prevent joint limit violations in advance.

The current operator hand pose is used as the desired goal pose for the avatar arm in the common gripper frame. We estimate the avatar arm joint configuration reaching this goal pose using inverse kinematics (IK). The latest received avatar arm joint positions are used to initialize the IK solver. The current joint velocities are approximated by a low-pass filtered version of the joint position first derivatives.
Having estimated the joint positions and velocities, we can apply the same avoidance strategy as described above (see \cref{eq:avoidance}).
Finally, the resulting joint torques can be transformed into the common 6D hand frame using the pseudoinverse of the Jacobian transpose $(J_A^T)^+$ and back to joint torques for the operator arm with $J_O^T$. This results in
\begin{equation}
 \tau_{la} = J_O^T (J_A^T)^+ \tau_{la-model}.
\end{equation}

The remaining two torque components from \cref{eq:operator_control} are $\tau_{no}$ and $\tau_{co}.$
The term $\tau_{no}$ is a null-space optimization term which pulls the elbow
towards a defined convenient pose in the null-space of the Jacobian. The result is a more human-like
elbow pose to maximize the operator workspace by pushing the arm away from singularities.
The last torque component is the Coriolis term $\tau_{co}$ obtained by the Panda model.

\subsubsection{Oscillation Observer}
\label{sec:observer}
Our telemanipulation control system as described above is designed as simple as possible,
but yet powerful to give the operator control over the robot along with necessary force feedback.
The downside is, that our controller cannot promise any stability, resulting in oscillations
when certain contact forces are measured. We address this issue by observing the force feedback channel,
detecting any critical oscillations, and reducing the feedback introduced into the system.

The sliding Discrete Fourier-Transformation (DFT) \citep{jacobsen2003sliding,jacobsen2004update} method is used
to analyze the force feedback channel per axis. Since we did not measure isolated oscillations in the torque channels,
analyzing the 3D forces is sufficient.
The force measurements are sampled with around 1\,kHz. We experimentally investigated
the oscillation frequency to obtain suitable DFT parameters and the observed frequency band.
In the experiments, we provoked vibrations in the system by placing a vase harshly on a table.
The used sliding DFT has resolution of 512 and uses the Hanning window to minimize spectral leakage.
We observe correlation between the generated oscillation and the amplitude of the
4th frequency (ca. 5.7\,Hz), is depicted in \cref{fig:dft_explain}.

In each time step, the force measurements of all three axes are analyzed using
separate sliding DFTs. Next, the combined result is computed using the Euclidean norm
over all axes amplitudes. The resulting value $v$ is clamped using experimentally obtained
parameters ($min = 163$ and $max = 500$) and scaled to be within the interval $[0,1]$.
Finally, the scalar $\beta = 1 - v$ is used to scale the force feedback provided to the operator (see \cref{sec:operator_arm_control}).
The change rate of $\beta$ is limited to gradually remove and restore the force feedback, s.t. the observer needs
0.8\,s to fully remove the feedback in case of detected oscillations and 1.7\,s to reach the normal force feedback control status again.
This gentle oscillation elimination behavior is as little noticeable for the operator as possible.
Note that \cref{fig:dft_explain} shows the oscillation without feedback reduction. The
computation of $\beta$ and the effect oscillation damping are evaluated in \cref{sec:dft_eval}.

The avatar arm controller commands the Panda arm of the avatar robot to follow the commanded 6D
pose by sending joint torques $\tau_a\in\R^7$ to the Franka Control Interface. The commanded torque is defined as
\begin{equation}
 \tau_a = \tau_{cmd} + \tau_{init} + \tau_{na} + \tau_{ca},
\end{equation}

where $\tau_{cmd}\in\R^7$ and $\tau_{init}\in\R^7$ are calculated to reach the goal pose during operation and initialization, respectively (see below).
The components $\tau_{na}$ and $\tau_{ca}$ are the null-space optimization and Coriolis terms similar to $\tau_{no}$ and $\tau_{co}$, as described in \cref{sec:control}.
The convenient elbow poses used for the null-space optimization is defined such that
the elbows are slightly stretched out. This generates a more human-like arm configuration
and keeps the arm away from singularities in the elbow joints. Other singularities do not occur due to the nature of the arm kinematic in the usable workspace and other joint position limits. Therefore, no special singularity handling is needed here.

The goal torques $\tau_{cmd}$ and $\tau_{init}$ are generated using a Cartesian impedance controller that emulates a spring--damper system. Its equilibrium point is the 6D goal pose commanded by the operator station in the common hand frame:
\begin{equation}
 \tau_{cmd} = J^T (-S \Delta p - D (J \dot{q})),
\end{equation}
where $J$ denotes the zero Jacobian, $S\in\R^{6\times6}$ and $D\in\R^{6\times6}$
denote the stiffness and damping matrix, $\Delta p \in\R^6$ is the error in translation
and rotation between the current and goal end-effector pose, and $\dot q\in\R^7$ denotes
the current joint velocities.
$\tau_{init}$ is only used for a safe initialization procedure (see below) and generated similarly using the current end-effector pose.
The stiffness and damping parameters symbols and their values are empirically tuned
to achieve some compliance while still closely following the operator command.

When no goal pose command is received, the controller keeps commanding the current arm pose to remain in a safe state.
This happens when the operator station is not active or if a communication breakdown occurs (no operator command within the last 100\,ms).
After receiving a command, the controller performs an initialization procedure which
fades linearly between the current and the new received goal pose. This prevents the robot
from generating high torques to suddenly reach the new, possibly distant pose.
This initialization process takes about 3\,s.

The Panda arm stops immediately when excessive forces are measured, for example when
there is unintended contact that exceeds force/torque thresholds. This feature is necessary to operate in a safe way.
After notification of the human operator, the avatar arm controller can restart
the arm automatically.
After performing the initialization procedure, normal teleoperation can be resumed.

\subsection{Hand Control}

The operator finger movements are captured using two SenseGlove haptic
interaction devices. Four separate finger
joint measurements are provided per finger. Since the Schunk SVH and SIH robotic hands
on the avatar have nine and five
actuated joints, respectively, only the corresponding joint measurements are selected and
linearly mapped to the
avatar hands. While this mapping does not precisely replicate hand
postures -- this is impossible anyways due to the
different kinematic structure -- it gives the operator full control over all hand DoFs.

Both hands provide feedback in the form of motor currents, which is used
to provide per-finger haptic feedback to the operator.
The SenseGlove brake system is switched on or off depending on a pre-defined
current threshold.

\subsection{Force-Torque Sensor Calibration}
\label{sec:ft_calibration}

Different end-effectors (SenseGloves, Schunk SIH, Schunk SVH hand,
and corresponding 3D printed mounting adapters) are
mounted on each of the four involved force-torque sensors. In addition,
sensor bias results in barely usable raw sensor data.
Thus, each sensor is calibrated separately to compensate these effects.
To this end, 20 data samples
from different sensor poses are collected. Each sample includes the
gravity vector in the sensor frame and the mean
of 100 sensor measurements from a static pose. A standard least squares
solver~\citep{4399184} is used to estimate the force-torque
sensor parameters, i.e., the force and torque bias and the mass and center of mass
of all attached components.
The same parameters including the additional mass and center of mass
transformation resulting by the force-torque
sensor itself is used to configure the built-in gravity compensation of
the Panda arms. The calibration is performed once after hardware changes
at the end-effectors or if the bias drift is too large.
This method does not compensate for
bias drift during usage, but is sufficient for our application.

\section{Evaluation}
\label{sec:eval}

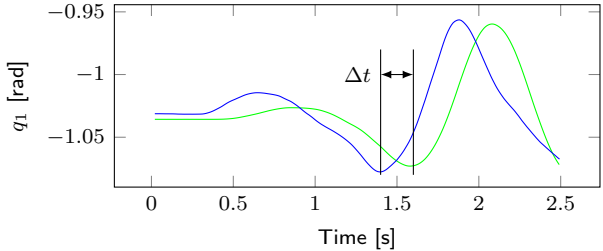
\begin{figure}[]
  \centering\begin{maybepreview}%
\pgfplotstableread{exp/feedback1.txt}\feedbacktable
\pgfplotstableread{exp/model1.txt}\modeltable
  \begin{tikzpicture}[font=\footnotesize\sffamily]
   \begin{axis}[height=4cm,width=.9\linewidth,
     xlabel={Time [s]}, ylabel={$q_1$ [rad]},]
     \addplot+ [mark=none, green, each nth point=10, filter discard warning=false, unbounded coords=discard] table [x=Time,y=j1] {\feedbacktable};
     \addplot+ [mark=none, blue, each nth point=10, filter discard warning=false, unbounded coords=discard] table [x=Time,y=j1] {\modeltable};
     \draw [black] (1.4, -1.08) -- ( 1.4, -0.98);
     \draw [black] (1.6, -1.08) -- ( 1.6, -0.98);
     \draw[latex-latex] (1.4,-1.0) node [anchor=east] {$\Delta t$} -- (1.6,-1.0) ;
   \end{axis}
  \end{tikzpicture}\end{maybepreview}%
  \caption{Predictive avatar model: Measured joint position for the first joint of the right avatar arm during a grasping motion (green) and predicted
  joint position for predictive limit avoidance (blue). Both measurements are captured on the operator side.
  Communication between both systems and motion execution generate a delay of up to 200\,ms ($\Delta t$), which
  is compensated by the predictive model.}
  \label{fig:exp_model}
\end{figure}

\begin{figure}[]
  \centering\begin{maybepreview}%
\pgfplotstableread{exp/ft2.txt}\fttable
\pgfplotstableread{exp/franka2.txt}\frankatable
  \begin{tikzpicture}[font=\footnotesize\sffamily]
   \begin{axis}[height=4cm,width=.9\linewidth,
     xlabel={Time [s]}, ylabel={Force [N]},
     ymin = -15, ymax =15,]
     \addplot+ [mark=none, blue] table [x expr=\coordindex,y=forceZ] {\fttable};
     \draw (40.0, 10.0) node [anchor=east, black] {Ours};
   \end{axis}
  \end{tikzpicture}\\
\begin{tikzpicture}[font=\footnotesize\sffamily]
   \begin{axis}[height=4cm,width=.9\linewidth,
     xlabel={Time [s]}, ylabel={Force [N]},
     ymin = -15, ymax =15,]
     \addplot+ [mark=none, blue] table [x expr=\coordindex,y=forceZ] {\frankatable};
     \draw (55.0, 10.0) node [anchor=east, black] {Panda};
   \end{axis}
  \end{tikzpicture}\end{maybepreview}%
  \caption{Operator arm movement: Force in z-direction (in the direction of the human palm)
  needed to move the arm in the same repetitive motion with our operator arm controller running (top)
  and using only the Panda built-in gravity compensation (bottom).}
  \label{fig:exp_ft}
\end{figure}
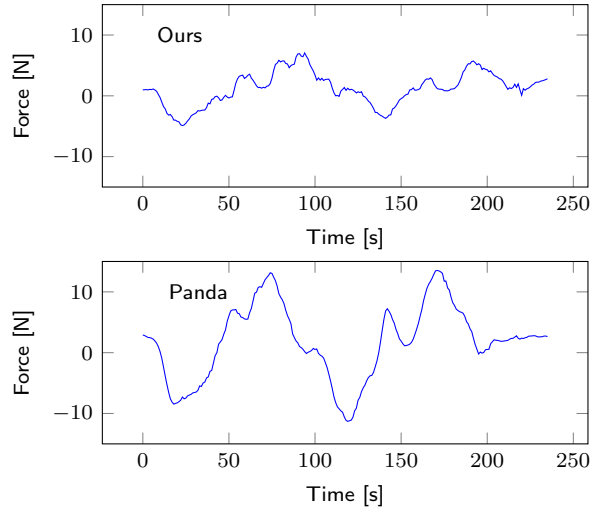

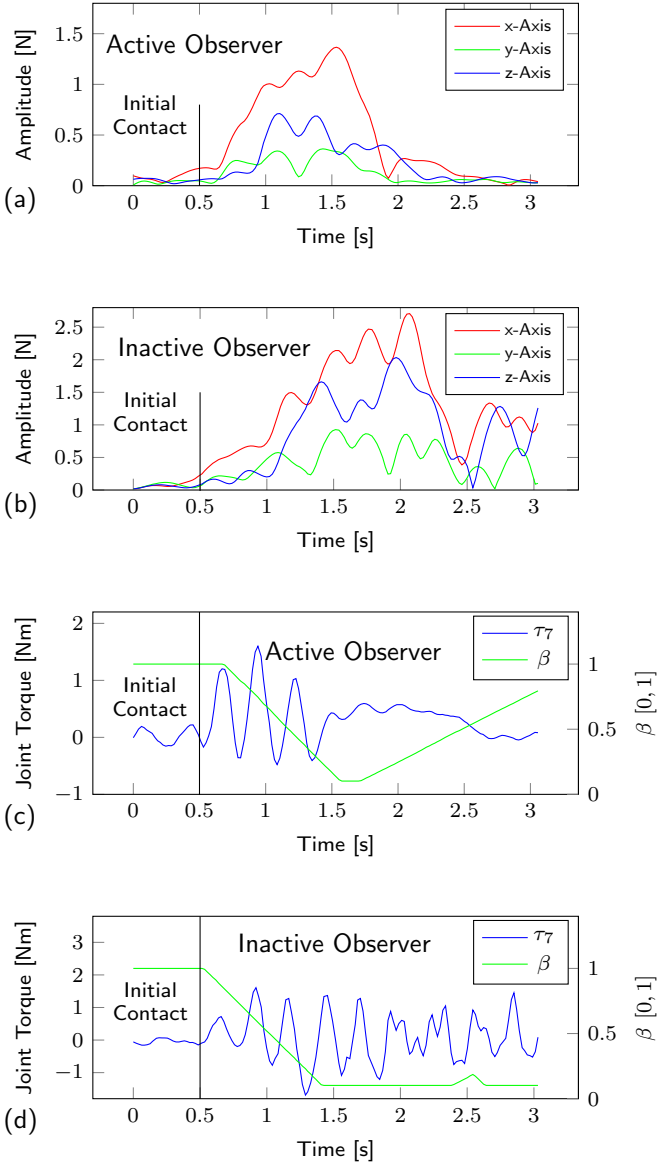
\begin{figure}[]
  \centering\begin{maybepreview}%
  \begin{tikzpicture}[font=\footnotesize\sffamily]
\def\xA{-1.5}
\def\yA{-1}
\def\xB{10.5}
\def\yB{3}

\pgfplotstableread{exp/dtf_exp_on.txt}\dtfexpontable
  \pgfresetboundingbox
     \path[use as bounding box] (\xA,\yA) rectangle (\xB,\yB);
   \begin{axis}[height=4cm,width=.9\linewidth,
     xlabel={Time [s]}, ylabel={Amplitude [N]},
     legend pos = north east,
     legend style={nodes={scale=0.8, transform shape}},
     ymin = 0, ymax = 1.8,
     ]

     \addplot+ [mark=none, red] table [x=t,y expr=\thisrowno{8} / 512] {\dtfexpontable};
     \addplot+ [mark=none, green] table [x=t,y expr=\thisrowno{9} / 512] {\dtfexpontable};
     \addplot+ [mark=none, blue] table [x=t,y expr=\thisrowno{10} / 512] {\dtfexpontable};
     \addlegendentry{x-Axis}
     \addlegendentry{y-Axis}
     \addlegendentry{z-Axis}
     \draw [black] (0.5, -1) -- (0.5, 0.8);
     \node[align=center, anchor=east, execute at begin node=\setlength{\baselineskip}{2.5ex}] at (0.48,0.7) {Initial\\Contact};
     \node[align=center, anchor=east] at (1.2,1.4) {\normalsize Active Observer};
   \end{axis}
     \node[align=center, anchor=west] at (-1.3,-0.2) {\normalsize (a)};
     \drawboundingbox[green]{\drawORnot}
  \end{tikzpicture}
\begin{tikzpicture}[font=\footnotesize\sffamily]
\def\xA{-1.5}
\def\yA{-1}
\def\xB{10.5}
\def\yB{3}

\pgfplotstableread{exp/dtf_exp_off.txt}\dtfexpofftable
     \path[use as bounding box] (\xA,\yA) rectangle (\xB,\yB);
   \begin{axis}[height=4cm,width=.9\linewidth,
     xlabel={Time [s]}, ylabel={Amplitude [N]},
     legend pos = north east,
     legend style={nodes={scale=0.8, transform shape}},
     ymin = 0, ymax = 2.8,
     ytick={0,  0.5,  1, 1.5, 2, 2.5},
     ]
     \addplot+ [mark=none, red] table [x=t,y expr=\thisrowno{8} / 512] {\dtfexpofftable};
     \addplot+ [mark=none, green] table [x=t,y expr=\thisrowno{9} / 512] {\dtfexpofftable};
     \addplot+ [mark=none, blue] table [x=t,y expr=\thisrowno{10} / 512] {\dtfexpofftable};
	\addlegendentry{x-Axis}
     \addlegendentry{y-Axis}
     \addlegendentry{z-Axis}
     \draw [black] (0.5, -1) -- (0.5, 1.5);
     \node[align=center, anchor=east, execute at begin node=\setlength{\baselineskip}{2.5ex}] at (0.48,1.2) {Initial\\Contact};
     \node[align=center, anchor=east] at (1.4,2.2) {\normalsize Inactive Observer};
   \end{axis}
   \node[align=center, anchor=west] at (-1.3,-0.2) {\normalsize (b)};
        \drawboundingbox[green]{\drawORnot}
  \end{tikzpicture}

  \begin{tikzpicture}[font=\footnotesize\sffamily]
  \def\xA{-1.5}
\def\yA{-1}
\def\xB{10.5}
\def\yB{3}

\pgfplotstableread{exp/dtf_exp_on.txt}\dtfexpontable
\pgfplotstableread{exp/dtf_exp_off.txt}\dtfexpofftable
     \path[use as bounding box] (\xA,\yA) rectangle (\xB,\yB);
  \begin{axis}[height=4cm,width=.9\linewidth,
     xlabel={Time [s]}, ylabel={Joint Torque [Nm]},
     legend pos = north east,
     legend style={nodes={scale=0.8, transform shape}},
     ylabel near ticks,
     ymin = -1, ymax = 2.2,
     xtick pos=left,
     ytick pos=left,
     ]
     \addplot+ [mark=none, blue] table [x=t,y=q7] {\dtfexpontable};
     \label{plot_1}
   \end{axis}
   \begin{axis}[height=4cm,width=.9\linewidth,
     ylabel={$\beta$ $[0,1]$},
     axis y line*=right,
     ylabel near ticks,
     hide x axis,
     ytick pos=right,
     ymin = 0, ymax = 1.4,
     ]
     \addlegendimage{/pgfplots/refstyle=plot_1}\addlegendentry{$\tau_7$}
     \addplot+ [mark=none, green] table [x=t,y=a] {\dtfexpontable};
     \addlegendentry{$\beta$}
     \draw [black] (0.5, -0.1) -- (0.5, 1.4);
     \node[align=center, anchor=east, execute at begin node=\setlength{\baselineskip}{2.5ex}] at (0.48,0.75) {Initial\\Contact};
     \node[align=center, anchor=east] at (2.4,1.1) {\normalsize Active Observer};
   \end{axis}
   \node[align=center, anchor=west] at (-1.3,-0.3) {\normalsize (c)};
        \drawboundingbox[green]{\drawORnot}
  \end{tikzpicture}

  \begin{tikzpicture}[font=\footnotesize\sffamily]
  \def\xA{-1.5}
\def\yA{-1}
\def\xB{10.5}
\def\yB{3}

\pgfplotstableread{exp/dtf_exp_off.txt}\dtfexpofftable
       \path[use as bounding box] (\xA,\yA) rectangle (\xB,\yB);
  \begin{axis}[height=4cm,width=.9\linewidth,
     xlabel={Time [s]}, ylabel={Joint Torque [Nm]},
     legend style={nodes={scale=0.8, transform shape}},
     legend pos = north east,
     ylabel near ticks,
     ymin = -1.8, ymax = 3.8,
     xtick pos=left,
     ytick pos=left,
     ytick={-1,  0,  1, 2,3},
     ]
     \addplot+ [mark=none, blue] table [x=t,y=q7] {\dtfexpofftable};
     \label{plot_2}
   \end{axis}
   \begin{axis}[height=4cm,width=.9\linewidth,
     ylabel={$\beta$ $[0,1]$},
     axis y line*=right,
     ylabel near ticks,
     hide x axis,
     ytick pos=right,
     ymin = 0, ymax = 1.4,
     ]
     \addlegendimage{/pgfplots/refstyle=plot_2}\addlegendentry{$\tau_7$}
     \addplot+ [mark=none, green] table [x=t,y=a] {\dtfexpofftable};
     \addlegendentry{$\beta$}
     \draw [black] (0.5, -0.1) -- (0.5, 1.4);
     \node[align=center, anchor=east, execute at begin node=\setlength{\baselineskip}{2.5ex}] at (0.48,0.75) {Initial\\Contact};
     \node[align=center, anchor=east] at (2.3,1.2) {\normalsize Inactive Observer};
   \end{axis}
   \node[align=center, anchor=west] at (-1.3,-0.3) {\normalsize (d)};
        \drawboundingbox[green]{\drawORnot}
  \end{tikzpicture}\end{maybepreview}%

  \caption{Oscillation observer module. Placing a vase onto a table results in the depicted amplitude
  response for 5.7\,Hz for each Cartesian axis ((a) and (b)).
  We performed the experiment twice with comparable executions.
  First, the oscillation observer was active ((a) and (c)). In the second execution, the observer was
  inactive ((b) and (d)).
  The corresponding joint torques commanded to the operator Panda arm are plotted exemplary for the 7th joint (c) and (d).
  The force feedback is scaled with $\beta$ (c) which is computed from the oscillation observer.
  Note that $\beta$ shown in (d) was computed but not used during execution.
  }
  \label{fig:exp_dft}
\end{figure}

In addition to our participation at the ANA Avatar XPRIZE Competition semifinals, we performed multiple experiments along with a small user study to evaluate the developed teleoperation system in our lab environment.

\subsection{Quantitative Experiments}
\label{sec:eval_workspace}

In a first experiment, we evaluated the operator arm workspace. 2,959 different 6D
left hand poses were captured from a sitting person performing typical arm motions
with a VR tracker on their wrist.
In addition to hand poses with a fully extended arm, most of the poses are directly
in front of the person, likely to be performed during manipulation tasks.
First, the initial arm mounting pose (motivated by the avatar configuration) of the operator arm was evaluated. Each captured hand pose was marked as reachable if an inverse kinematic solution for the arm was found. In a second step, different arm mounting poses were sampled to find an optimal pose, maximizing the number of reachable hand poses (see \cref{fig:exp_arm_reach}). \cref{tab:arm_workspace} reports quantitative results.
The resulting arm mounting pose drastically increases the overlap (from 60.6\% to 96.3\%) between the human operator's and the avatar's arm workspace, but requires
a more complicated mounting setup.

\begin{table}
 \centering
 \caption{Operator arm workspace analysis}\label{tab:arm_workspace}
 \begin{tabular}{lrrr}
  \toprule
  Mounting Pose & Reached & Missed & Reached [\%]\\
  \midrule
  Initial   & 1,795 & 1,164 & 60.6\,\%\\
  Optimized & 2,848 & 111 & 96.3\,\%\\
  \bottomrule
 \end{tabular}
\end{table}

In a second experiment, we evaluated the predictive limit avoidance module. Avatar arm position and velocity limits are haptically displayed via joint forces to the operator. Since measured joint positions and velocities are afflicted with latency generated by network communication ($<$1\,ms) and motion execution using the Cartesian impedance controller, which can reach up to 200\,ms (see \cref{sec:avatar_arm_control}),
the operator control predicts the avatar arm joint configuration. \cref{fig:exp_model} shows the measured and predicted joint position of the first right arm joint. The prediction
compensates the delays, which allows for instantaneous feedback of the avatar arm limits to the operator.

In a third experiment, we investigated the forces and torques required to move the operator station arm, since this
directly affects operator fatigue.
We measured the forces and torques applied to the arm by reading the force-torque sensor measurements. The arm was
moved in a comparable manner once with only the Panda gravity compensation enabled (i.e., $\tau_{cmd} = 0$ see \cref{eq:operator_control}) and a second time with our
arm force controller running. In \cref{fig:exp_ft}, the forces in the direction of one exemplary axis are shown.
The results demonstrate the advantage of using an external force-torque sensor to generate a more unencumbered feeling for the operator
while using the system.

\label{sec:dft_eval}
In the last experiment, we analyzed the oscillation observer module (see \cref{sec:observer}), which observes the avatar force feedback channel
to detect and prevent oscillations in the closed-loop controller. We used the avatar system to place a metal vase harshly on a table (see \cref{fig:dft_explain}).
The executed joint torques on the operator side along with the frequency responses of the force-feedback measurements are shown in \cref{fig:exp_dft}.
Although the oscillations are not eliminated immediately (c), reducing the feedback gain yields the expected oscillation suppression.
The oscillation is detected by analyzing the frequency response shown in (a) and (b). The gain $\beta$ is reduced with the maximum allowed rate to a value of 0.1 within 0.8\,s.
Next, the operator does not feel the contact forces as strong, the oscillation stops (shown in the exemplary
torque command for the 7th operator arm joint (c)), and the gain is increased again, which brings back
the force feedback for the operator.
In conclusion, the oscillation observer module is a necessary part of our
simple force-feedback control loop to ensure safe operation.

\subsection{User Study}
\label{sec:user_study}
Our goal was to create an immersive and intuitive feeling for operation at remote locations using our system.
Since humans have their very own preferences and subjective feelings of how good or intuitive certain control mechanisms
perform, we carried out a user study with untrained operators, comparing different telemanipulation approaches.
Due to the COVID-19 pandemic, we were limited to immediate colleagues as subjects, which severely constrained
the scope of our study. Although all participants had a rough idea of the system, they controlled it for the first time during this study.

\begin{figure}
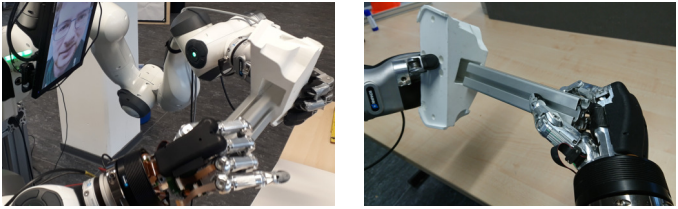

\begin{maybepreview}%
 \includegraphics[height=2.7cm]{images/experiments/task2.jpg}
 \hfill
 \includegraphics[height=2.7cm]{images/experiments/task3.jpg}
 \end{maybepreview}%
 \caption{User study with untrained operator: Both objects had to be grasped and the bar had to be inserted into the hole.}
 \label{fig:exp_task}
\end{figure}

A total of five participants were asked to perform a bimanual peg-in-hole manipulation task. First, two different objects had to be grasped: a small aluminum
bar and a 3D printed part with a hole. Afterwards, the bar should be inserted into the hole (see~\cref{fig:exp_task}). The avatar robot was already
placed in front of a table and both objects were within the avatar's workspace. Participants controlled the robot using the operator station located within the same room.
Only the HMD with the avatar's perspective was used for visual feedback.
The task was challenging due to very little friction between the finger and objects and tight tolerances, which required precise insertion alignment.

Each participant performed the task three times with the following control modes:
\begin{enumerate}
 \item Operator station with force feedback enabled,
 \item Operator station with force feedback disabled, and
 \item VR controllers.
\end{enumerate}

In the first control mode, all system components were active as described in this article. Any force and haptic feedback were disabled for the second control mode, i.e.
the operator had to rely on visual feedback only.
In the third control mode two HTC Vive VR controllers were used as input devices. As long as the trigger
button was pressed, the corresponding avatar arm followed the controller movement.
A different button was programmed to toggle between a defined closed and open hand pose.

A maximum of 5\,min were granted to solve the task before it was marked as a failure. An object dropped
outside the reachable workspace resulted in a failed trial. Objects dropped onto the table within the workspace
of the avatar could be grasped again with no penalty.
The participants were allowed to test each control mode about 1\,min before starting the measured test.

\begin{figure*}[]
\scalebox{0.9}{
\begin{maybepreview}%
 ﻿\begin{tikzpicture}[font = \footnotesize, every mark/.append style={mark size=0.5pt}]
 \begin{axis}[
     name=plot,
     boxplot/draw direction=x,
     width=0.6\textwidth,
     height=7.5cm,
     boxplot={
         draw position={1.5 - 0.325/3 + 1.0*floor((\plotnumofactualtype + 0.001)/3) + 0.2*mod((\plotnumofactualtype + 0.001),3)},
         box extend=0.17,
         average=auto,
         every average/.style={/tikz/mark=x, mark size=1.5, mark options=black},
         every box/.style={draw, line width=0.5pt, fill=.!40!white},
         every median/.style={line width=2.0pt},
         every whisker/.style={dashed},
     },
     ymin=1,
     ymax=10,
     y dir=reverse,
     ytick={1,2,...,11},
     y tick label as interval,
     yticklabels={
Did you feel safe and comfortable?,
Did you feel like you were handling the objects directly?,
Was it easy to control the robot?,
Was it intuitive to control the arms?,
Was it intuitive to control the fingers?,
Did you find and recognize the objects?,
Was it easy to grasp the objects?,
Was it easy to fit the objects together?
     },
     y tick label style={
         align=center
     },
     xmin=0.75,
     xmax=7.25,
     xtick={1, 2 ,..., 7},
     xticklabels = {1, 2, ..., 7},
     cycle list={{green!50!black,orange!50!red,blue}},
     y dir=reverse,
     legend image code/.code={
         \draw [#1, fill=.!40!white] (0cm,-1.5pt) rectangle (0.3cm,1.5pt);
     },
     legend style={
         anchor=north west,
         at={($(0.0,1.0)+(0.2cm,-0.1cm)$)},
     },
     legend cell align={left},
 ]

 \addplot
 table[row sep=\\,y index=0] {
 data\\
 7\\7\\7\\6\\6\\
 };

 \addplot
 table[row sep=\\,y index=0] {
 data\\
 7\\7\\4\\6\\7\\
 };

 \addplot
 table[row sep=\\,y index=0] {
 data\\
 7\\7\\5\\6\\7\\
 };

 \addplot
 table[row sep=\\,y index=0] {
 data\\
 3\\1\\3\\3\\5\\
 };

 \addplot
 table[row sep=\\,y index=0] {
 data\\
 6\\4\\6\\5\\7\\
 };

 \addplot
 table[row sep=\\,y index=0] {
 data\\
 5\\6\\5\\5\\7\\
 };

 \addplot
 table[row sep=\\,y index=0] {
 data\\
 6\\2\\6\\4\\6\\
 };

 \addplot
 table[row sep=\\,y index=0] {
 data\\
 4\\5\\4\\6\\6\\
 };

 \addplot
 table[row sep=\\,y index=0] {
 data\\
 4\\7\\5\\5\\7\\
 };

 \addplot
 table[row sep=\\,y index=0] {
 data\\
 6\\4\\6\\5\\5\\
 };

 \addplot
 table[row sep=\\,y index=0] {
 data\\
 5\\5\\6\\6\\6\\
 };

 \addplot
 table[row sep=\\,y index=0] {
 data\\
 4\\7\\5\\6\\6\\
 };

 \addplot
 table[row sep=\\,y index=0] {
 data\\
 5\\2\\1\\5\\4\\
 };

 \addplot
 table[row sep=\\,y index=0] {
 data\\
 5\\7\\3\\6\\6\\
 };

 \addplot
 table[row sep=\\,y index=0] {
 data\\
 4\\7\\4\\6\\6\\
 };

 \addplot
 table[row sep=\\,y index=0] {
 data\\
 7\\7\\7\\6\\7\\
 };

 \addplot
 table[row sep=\\,y index=0] {
 data\\
 7\\7\\7\\6\\7\\
 };

 \addplot
 table[row sep=\\,y index=0] {
 data\\
 7\\7\\7\\6\\7\\
 };

 \addplot
 table[row sep=\\,y index=0] {
 data\\
 5\\4\\7\\2\\6\\
 };

 \addplot
 table[row sep=\\,y index=0] {
 data\\
 4\\5\\4\\5\\6\\
 };

 \addplot
 table[row sep=\\,y index=0] {
 data\\
 2\\7\\6\\5\\7\\
 };

 \addplot
 table[row sep=\\,y index=0] {
 data\\
 5\\2\\5\\2\\2\\
 };

 \addplot
 table[row sep=\\,y index=0] {
 data\\
 3\\4\\4\\6\\6\\
 };

 \addplot
 table[row sep=\\,y index=0] {
 data\\
 1\\7\\4\\6\\6\\
 };

 \legend{VR, Without FF, With FF}

 \end{axis}

 \draw[-latex] ($(plot.south west)+(0.3cm,0.2cm)$) -- ($(plot.south west)+(1.4cm,0.2cm)$)  node[midway,above,font=\scriptsize,inner sep=1pt] {better};

 \end{tikzpicture}
 \end{maybepreview}%

}%
\caption{Qualitative results of our user questionnaire. We show the median, lower and upper quartile (includes interquartile range), lower and upper fence, outliers (marked with •) as well as the average value (marked with $\times$), for each aspect as recorded in our questionnaire.}
\label{fig:user_study}
\end{figure*}
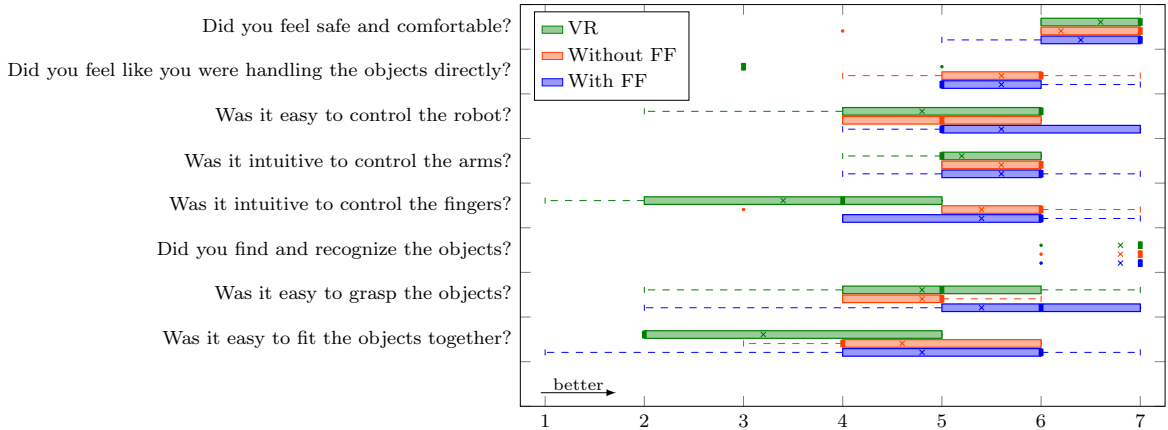

\begin{table}
 \centering
 \footnotesize
 \caption{User study success rates and timings.}
 \label{tab:success}
 \begin{tabular}{lrrr}
  \toprule
  Telemanipulation mode & Success & \multicolumn{2}{c}{Completion time [s]} \\
  \cmidrule (lr) {3-4}
              &         & \hspace{2em}Mean & StdDev \\
  \midrule
  1) Exoskeleton with feedback   & 4/5 & 119.0 & 117.1\\
  2) Exoskeleton w/o feedback   & 5/5 & 123.0 & 88.4\\
  3) VR controllers           & 3/5 & 126.3 & 25.8\\
  \bottomrule
 \end{tabular}
\end{table}

\cref{tab:success} reports the quantitative results of the user study. The time needed to successfully solve
the task is quite similar over the different telemanipulation modes. From these experiments, we
realized that the completion time was highly influenced by external factors, such as losing the object due to not enough
finger friction or different grasping and object handling solutions, which are unrelated to the used operator interface. In addition,
humans can easily compensate missing force feedback using visual feedback. All three unsuccessful trials failed due to reaching the
maximum experiment time of 5\,min.
To generate a more meaningful statement based on performance scores, more test samples are needed.

In addition to these quantitative measurements, we asked the participants to
answer a short questionnaire about each telemanipulation mode with answers
from the 1-7 Likert scale (see \cref{fig:user_study}).
The results show that especially the feeling of handling the objects and
intuitive finger control was subjectively much better using
the Operator Station. Enabling the force and haptic feedback gives the highest
advantage when picking up the objects from the table.
This can be explained by the additional feedback indicating contact between
the hand and the table which cannot be perceived visually due to occlusions.
All participants reported to feel safe and comfortable using the system.
Although the experiment time was limited, this suggests non-excessive cognitive load
on the operator.

Overall, the user study showed that our developed system is intuitive to use for
untrained operators. Even though the force and haptic feedback did not increase the success rate
of solving the task, it increases the immersive feeling as shown by the questionnaire.

\subsection{ANA Avatar XPRIZE Competition}
\label{sec:eval_xprize}

\begin{figure}
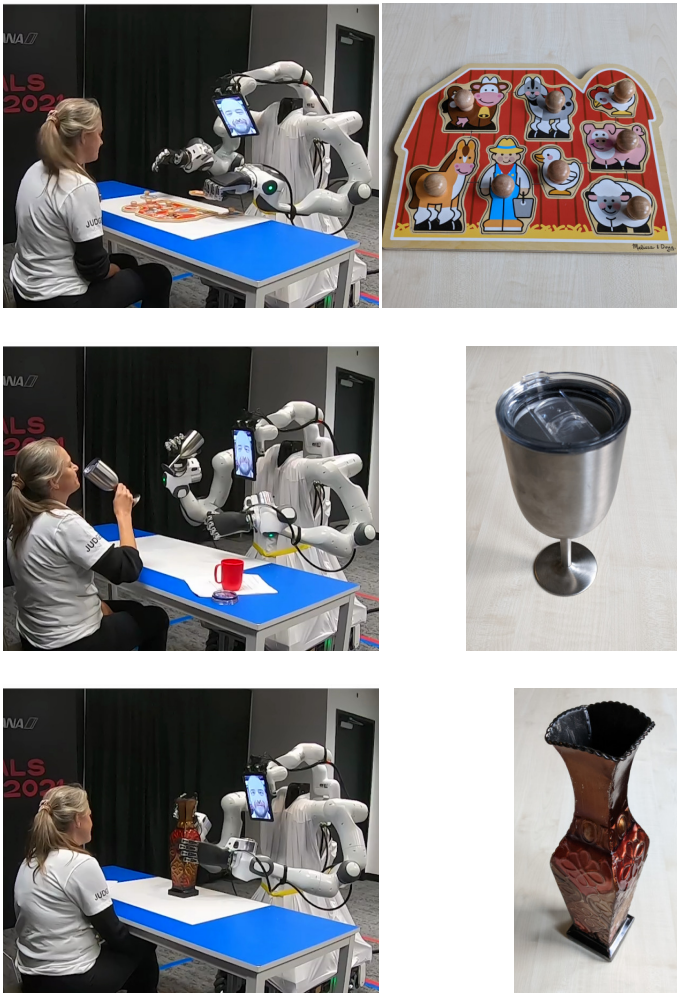

\begin{maybepreview}%
 \includegraphics[height=4.08cm, trim={1100 300 120 200},clip]{images/competition/puzzle.png}%
 \hfill%
 \includegraphics[height=4.03cm, trim={100 550 100 550},clip]{images/competition/puzzle.jpg}\\

 \includegraphics[height=4.08cm, trim={1100 300 120 200},clip]{images/competition/business.png}%
 \hfill%
 \includegraphics[height=4.03cm, trim={400 700 400 100},clip]{images/competition/glas.jpg}\\

 \includegraphics[height=4.08cm, trim={1100 300 120 200},clip]{images/competition/museum.png}%
 \hfill%
 \includegraphics[height=4.03cm, trim={500 500 700 100},clip]{images/competition/vase.jpg}\\
 \end{maybepreview}%
 \caption{ANA Avatar XPRIZE Semifinal Scenarios (left) and the used objects (right). The three scenarios were: Solving a jigsaw puzzle (top), celebrating a business deal (middle), and exploring an artifact (bottom).}
 \label{fig:semifinal_scenarios}
\end{figure}

\begin{table}
 \centering
 \caption{Avatar Arm safety stops}\label{tab:safety}
 \begin{tabular}{llll}
  \toprule
  & Scenario 1 & Scenario 2 & Scenario 3\\
  \midrule
  Day 1 & Exceeded& Exceeded & None\\
  &  torque limits & vel. limits & \\
  Day 2 & None & None & Software failure \\
  \bottomrule
 \end{tabular}
\end{table}

\begin{figure*}[]
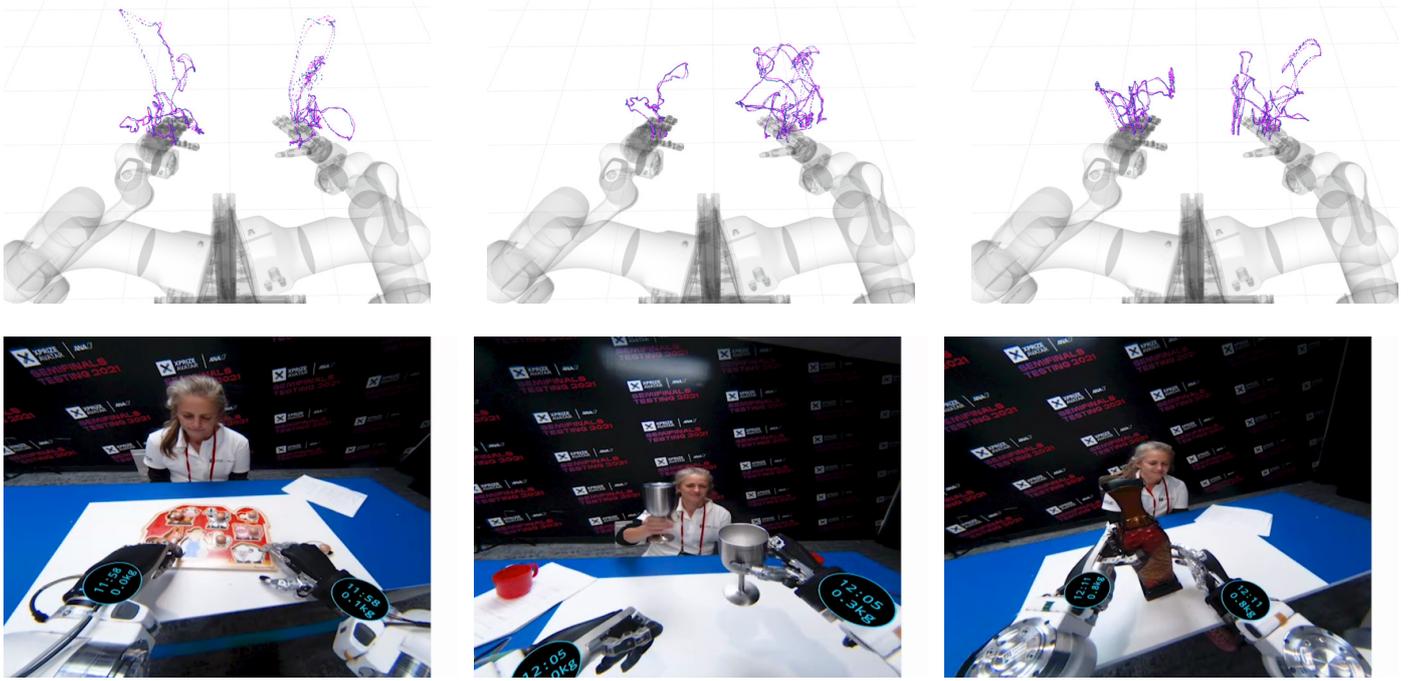

 \begin{maybepreview}%
 \includegraphics[height=4.0cm, trim={570 250 370 100},clip]{images/competition/workspace_puzzle.png}
 \hfill
 \includegraphics[height=4.0cm, trim={570 250 370 100},clip]{images/competition/workspace_business.png}
 \hfill
 \includegraphics[height=4.0cm, trim={570 250 370 100},clip]{images/competition/workspace_museum.png}\\
 
 \includegraphics[height=4.5cm, trim={100 600 1250 50},clip]{images/competition/vr_view_puzzle.png}
 \hfill
 \includegraphics[height=4.5cm, trim={100 600 1250 50},clip]{images/competition/vr_view_business.png}
 \hfill
 \includegraphics[height=4.5cm, trim={100 600 1250 50},clip]{images/competition/vr_view_museum.png}%
 \end{maybepreview}%
 \caption{Workspace analysis for all scenarios (solving a puzzle, celebrating a business deal, exploring an artifact) during the ANA Avatar XPRIZE Competition semifinal. Top: Hand positions of the avatar (blue) and operator (magenta) depicted in the common frame. Bottom: Operator VR perspective.}
 \label{fig:workspace_semi}
\end{figure*}

Our avatar system was evaluated by independent judges during the ANA Avatar XPRIZE competition semifinals over two days\footnote{Video material about the competition can be found here: \url{http://ais.uni-bonn.de/nimbro/AVATAR/}}. At each competition day, a first judge acted as the operator who
performed 18 tasks in three predefined scenarios together with the second judge acting as the so called ``recipient''. The recipient was sitting with the avatar robot at a table
in a different room about 100\,m away from the operator control room. Communication of any kind between both judges was only possible through the avatar system.
Both judges were trained to get comfortable with our system during the first hour of a trial.
In the second hour, the judges had to solve the tasks without any instructions or support from our team. Thus, the judges were
no experts but slightly more trained compared to the completely untrained operators in our user study (see \cref{sec:user_study}).
\cref{fig:semifinal_scenarios} shows the three scenarios: Solving a jigsaw puzzle, celebrating a business deal, and exploring an artifact from a historical exhibition with the robot's senses. The enabled force feedback (Control Mode 1, see \cref{sec:user_study}) enabled the operator to feel some texture of the artifact.
The judges evaluated the avatar system with a major focus on the ability to convey human senses, actions, and presence in the remote location in real time.
At both competition days, the same three scenarios were tested by different judges. The better score per scenario (max. 30 points per scenario) counted towards the final score.
Additional 10 points were given based on a video submitted prior to the semifinal showing the system in action in our lab\footnote{\url{https://www.youtube.com/watch?v=yGwJIDBMolk}}.
Our team NimbRo archived an almost perfect score of 99 out of 100 points, which placed us first in the semifinal.

We analyzed the recorded data during the official semifinal test runs to obtain more insight on the technical performance of our system.
In the following, we report some results.

\subsubsection{Safety System}
\label{sec:safety_eval}
One important aspect of our avatar robot is to operate safely next to and in cooperation with humans. Therefore, the avatar arm controller stops the Panda arm immediately when unexpected high forces/torques are measured (see \cref{sec:avatar_arm_control}). This behavior was activated three times during the performance of the six official scenarios (see \cref{tab:safety}).

\pgfplotstableread{exp/delay_means.txt}\delaytable
\begin{figure}[]
  \centering\begin{maybepreview}%
  \begin{tikzpicture}[font=\footnotesize\sffamily]
   \begin{axis}[height=5cm,width=.95\linewidth,
     xlabel={Temporal Offset [ms]}, ylabel={Translation Error [mm]},]
     \addplot+ [mark=none, blue, each nth point=1, filter discard warning=false, unbounded coords=discard] table [x=Delay,y=Error] {\delaytable};
   \end{axis}
  \end{tikzpicture}
  \end{maybepreview}%
  \caption{Avatar arm latency analysis. The graph shows the resulting mean translation error, comparing the current avatar hand position with an operator command shifted into the future. The mean is calculated over both arms and all six official competition scenarios.
  The minimum can be found at 44\,ms, which is the estimated round-trip latency in our system --- including network and motion execution.}
  \label{fig:exp_delay}
\end{figure}
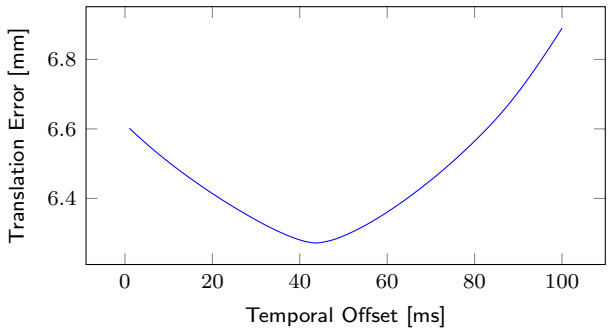

At Day 1, both arm stops were triggered by exceeded torque and velocity limits. In the first case, operator and recipient performed a powerful 'high five' gesture,
which resulted in a rapid increase of torque applied to the robot, exceeding the joint torque limits. The predictive limit avoidance module was not able
to prevent the operator executing such high contact forces, since the operator arm torque limits were reached as well.
In the second scenario on Day 1, the operator waved to the recipient. The different kinematic chains of the operator and avatar arm
demand high joint accelerations and speeds on the avatar side for relative low operator arm speeds and accelerations. The limit predictive module tried to slow down the
operator waving speed, but humans can overcome these applied feedback forces to always give the operator control for safety reasons. Here, the operator learned immediately continuing waving in a slightly slower and less acceleration-needing manner.
At the second competition day, the right avatar arm stopped once while not actively being moved. After some investigation, we
found a software failure which led to a very unlikely race condition. This was easily fixed after the competition.

In all cases, the avatar and operator robot continued to work in a safe manner, which is necessary for human-robot interactions.
Our intuitive feedback to the operator gave instantaneous situation awareness and allowed the operator to react and learn immediately
to continue the ongoing tasks.

\subsubsection{Arm Controller Accuracy}
\label{sec:arm_controller_accuracy}

After optimizing the usable workspace for the operator (see \cref{sec:eval_workspace}), \cref{fig:workspace_semi} shows the used
workspace for all three scenarios at the second competition day. The required workspaces during the competition tasks were relatively low, such
that we had no issues with our system.
We used the same data to analyze the spatial and temporal accuracy of our arm controller,
i.e., how precise and fast the avatar arms follow the operator commands during an evaluation task.

The measured operator and avatar hand positions were captured with 1000\,Hz over the duration of the six official
competition scenarios. We then calculated the translation error between the corresponding target and goal position for each arm,
resulting in a 6.6\,mm mean translation error.
In addition, we investigated the estimated delay in our arm controller including
network communication, controller runtime, and motion execution. We compared the measured
hand position of the avatar with the up to 100\,ms delayed operator's arm control position
and calculated the mean translation error as explained above.
\cref{fig:exp_delay} shows the mean translation error over all competition tasks and both
arms for a given temporal offset. The minimum translation error is archived
using a temporal shift of 44\,ms, resulting in a mean translation error of 6.3\,mm.
The results with and without temporal shift for both arms and all tasks are reported in \cref{tab:semi_accuracy}.
Noticeable is the comparably large error for the right arm during Scenario 1 and 2
at the first competition day. This can be explained by the executed safety stops (see \cref{sec:safety_eval}).
It takes about two seconds to safely fade the avatar arm to the commanded pose. This results in possibly
large position errors.
Overall, the observed errors are rather small and not noticeable for the operator and could potentially be
decreased by an even more accurate calibration on a whole-system level. The estimated round-trip execution latency of 44\,ms
could be considered for future system improvement.

\begin{table}
\centering
\caption{Translation error [mm] at ANA Avatar XPRIZE semifinals.}
\begin{threeparttable}
 \label{tab:semi_accuracy}
 \begin{tabular}{lrrrrrr}
  \toprule
  & \multicolumn{3}{c}{Day 1} & \multicolumn{3}{c}{Day 2}\\
  Scenario & 1 & 2 &  3 &  1 &  2 &  3 \\
  \midrule
  Left Arm          & 6.4 & 5.7 & 7.9 & 4.9 & 3.8 & 4.8\\
  44\,ms shift\tnote{1}    & 5.8 & 5.3 & 7.6 & 4.6 & 3.6 & 4.5\\
  \midrule

  Right Arm         & 13.2 & 9.5 & 6.4 & 6.3 & 5.5 & 5.1\\
  44\,ms shift\tnote{1}      & 12.5 & 9.2 & 6.0 & 6.3 & 5.2 & 4.8\\
  \bottomrule
 \end{tabular}
 \begin{tablenotes}
\item[1]\footnotesize{Delay between operator station and avatar (see \cref{fig:exp_delay})}
\end{tablenotes}
\end{threeparttable}
\end{table}

\section{Discussion \& Conclusion}

This work presented a bimanual telemanipulation system consisting of an exoskeleton-based operator control station and an anthropomorphic avatar robot.
Both components communicate using our force and haptic feedback controller, which allows safe and intuitive teleoperation
for both the operator and persons directly interacting with the avatar.
The control method is agnostic to the kinematic parameters and uses only a common Cartesian hand frame for
commands and feedback. Using the predictive limit avoidance avatar model, arm limits for both the operator and avatar side
can be force-displayed to the operator with low latency. The oscillation observer and damper modules detect and suppress
oscillations in the feedback control loop by reducing the force feedback gains temporarily.
Additional force-torque sensors measurements are used to generate a weightless feeling for the operator while
moving the arms without establishing contact on the avatar side.

We evaluated the system using a user study with untrained operators
as well as in lab experiments. In addition, the system performed very well at the ANA Avatar XPRIZE Competition Semifinals,
scoring 99 out of 100 points. This demonstrates the intuitiveness and
reliability of our system and its control methods.

\newpage
\section*{Acknowledgments}
This work has been funded by the Deutsche Forschungsgemeinschaft (DFG, German Research Foundation) under Germany’s Excellence Strategy,\\
EXC-2070 - 390732324 - PhenoRob.

\bibliography{references}

\end{document}